\documentclass[lettersize,journal]{IEEEtran}
\usepackage{amsmath,amsfonts}
\usepackage{array}
\usepackage[caption=false,font=normalsize,labelfont=sf,textfont=sf]{subfig}
\usepackage{textcomp}
\usepackage{stfloats}
\usepackage{url}
\usepackage{verbatim}
\usepackage{graphicx}
\usepackage{subcaption}

\usepackage{subfig} 
\usepackage[margin=0.5in]{geometry}

\usepackage[caption=false,font=normalsize,labelfont=sf,textfont=sf]{subfig}
\usepackage{multirow}

\usepackage{bm}
\usepackage{color}
\usepackage{nomencl}
\usepackage[linesnumbered,ruled,vlined]{algorithm2e}
\usepackage[hidelinks]{hyperref}

\usepackage{amsthm}
\usepackage{amssymb}
\usepackage[noend]{algpseudocode}

\usepackage{amsmath}
\usepackage{float}
\usepackage{tabularx}
\usepackage{booktabs}
\usepackage{tikz}
\usepackage{bm}
\usepackage{nomencl}
\makenomenclature

\def\BibTeX{{\rm B\kern-.05em{\sc i\kern-.025em b}\kern-.08em
    T\kern-.1667em\lower.7ex\hbox{E}\kern-.125emX}}
\usepackage{balance}
\begin{document}
\title{Just in time Informed Trees:  Manipulability-Aware Asymptotically Optimized Motion Planning}
\author{Kuanqi Cai*$^{1}$, Liding Zhang*$^{2}$, Xinwen Su$^{2}$, Kejia Chen$^{2}$, Chaoqun Wang$^{3}$, Sami Haddadin$^{4}$, ~\IEEEmembership{Fellow,~IEEE}, \\ Alois Knoll$^{2}$, ~\IEEEmembership{Fellow,~IEEE}, Arash Ajoudani\textsuperscript{+}$^{1}$, Luis Figueredo\textsuperscript{+}$^{5}$  
\thanks{$^{1}$K. Cai and A. Ajoudani are with the Human-Robot Interfaces and Interaction(HRI2), Istituto Italiano Di Tecnologia (IIT), Genova, 16163, Italy. K. Cai is also with the Swiss Federal Technology Institute of Lausanne (EPFL).}
\thanks{$^{2}$L. Zhang, X. Sun, K. Chen, S. Haddadin, and A. Knoll are with the Technical University of Munich, Munich, 85748, Germany. }
\thanks{$^{3}$C. Wang is with the School of Control Science and Engineering, Shandong University, Shandong, 250100, China.}
\thanks{$^{4}$S. Haddadin is the Vice President of Resarch at Mohamed bin Zayed University of AI, Abu Dhabi.}
\thanks{$^{5}$L. Figueredo is with the School of Computer Science, University of Nottingham, UK. He is also an Associate Fellow at the MIRMI, Techn. Univ. of Munich.}
\thanks{$^{*}$Equal contribution, \textsuperscript{+} Equal supervision.} \thanks{Corresponding Author: Chaoqun Wang}
\thanks{This work was supported in part by the European Union Horizon Project TORNADO (GA 101189557), and by the Lighthouse Initiative Geriatronics -- StMWi Bayern (Project X, 5140951), LongLeif GaPa gGmbH (Project Y, 5140953). The authors would like to thank the KI.FABRIK Bayern (grant DIK0249) project.}
}

\maketitle

\begin{abstract}
In high-dimensional robotic path planning, traditional sampling-based methods often struggle to efficiently identify both feasible and optimal paths in complex, multi-obstacle environments. This challenge is intensified in robotic manipulators, where the risk of kinematic singularities and self-collisions further complicates motion efficiency and safety. To address these issues, we introduce the Just-in-Time Informed Trees (JIT*) algorithm, an enhancement over Effort Informed Trees (EIT*), designed to improve path planning through two core modules: the Just-in-Time module and the Motion Performance module. The Just-in-Time module includes "Just-in-Time Edge," which dynamically refines edge connectivity, and "Just-in-Time Sample," which adjusts sampling density in bottleneck areas to enable faster initial path discovery. The Motion Performance module balances manipulability and trajectory cost through dynamic switching, optimizing motion control while reducing the risk of singularities.
Comparative analysis shows that JIT* consistently outperforms traditional sampling-based planners across $\mathbb{R}^4$ to $\mathbb{R}^{16}$ dimensions. Its effectiveness is further demonstrated in single-arm and dual-arm manipulation tasks, with experimental results available in a video at \href{https://youtu.be/nL1BMHpMR7c}{\textcolor{blue}{https://youtu.be/nL1BMHpMR7c}}.

\end{abstract}

\begin{IEEEkeywords}
Sampling-based path planning, manipulable, optimal planning, collision avoidance

\end{IEEEkeywords}

\section{Introduction}

Path planning is a fundamental problem in robotics, centered on finding an optimal, collision-free path from start to target while adhering to system-specific motion constraints \cite{caiiros}. In practice, this task can be computationally intensive, particularly in high-dimensional spaces while optimizing motion performance. Various methods have been developed to address these challenges, with sampling-based algorithms proving especially effective for high-dimensional, complex environments. Techniques such as Rapidly-exploring Random Tree (RRT) \cite{lavalle1998rapidly} and Probabilistic Roadmap (PRM) \cite{kavraki1996probabilistic} achieve obstacle avoidance by randomly sampling points to construct feasible paths. However, these methods often require significant time to find initial solutions and converge to optimal paths, especially in narrow free-space settings. To improve efficiency, the RRT* and PRM* methods enhance path optimality and convergence by incorporating rewiring and adaptive neighbor connection techniques. \textcolor{black}{However, classical nearest neighbor methods, such as $k$-NN~\cite{Xue2004} and $r$-NN~\cite{1961r_near}, optimize connections between sampled points but rely on fixed parameters $k$ and $r$, which limits their adaptability across diverse scenarios. Moreover, neighboring and rewiring are constrained to the parent node of the tree, leading to suboptimal paths and restricting the ability to efficiently find low-cost initial solutions.}

Sample-based optimization enhances planning efficiency by refining sample generation~\cite{caicurious}. Traditional methods use uniform random sampling~\cite{webb2013kinodynamic} to explore the space but converge slowly due to high randomness. Smart~\cite{islam2012rrt} and informed sampling~\cite{gammell2014informed} improve efficiency but struggle in dense obstacles and narrow passages. These methods often discard and resample upon collision, increasing time consumption, especially in high-dimensional planning.
Existing sampling methods are unsuitable for robotic manipulators, as the shortest path may cause kinematic singularities or self-collisions, leading to infeasible joint velocities and restricted motion. Incorporating manipulability is crucial to avoid these issues. While studies such as Shen~\cite{shen2023adaptive}, Kaden~\cite{kaden2019maximizing}, and Zhang~\cite{zhang2024path} integrate manipulability into RRT*, their end-effector-focused approaches limit applicability in high-dimensional, multi-arm scenarios. Moreover, planning in operational space must account for manipulator-specific constraints like joint limits, increasing complexity, and slowing convergence to optimal solutions.

To efficiently solve high-dimensional planning problems with consideration for robotic arm performance, we propose Just-in-Time Informed Trees (JIT)*, an anytime optimal planner. JIT combines two key modules: Just-in-Time, which adaptively forms edges and densifies sampling in critical regions to enhance search efficiency, and Motion Performance, which balances path cost and manipulability to dynamically optimize motion beyond shortest-path solutions.

The contributions of this work are summarized as follows: 
\begin{itemize} 
\item Introduce the Just-in-Time Module (Edge and Sample), which dynamically expands edges and optimizes sampling in critical regions to improve path planning efficiency.
\item Integrate motion performance optimization by balancing path cost and manipulability while accounting for self-collision checks to achieve an optimal solution. 
\item Demonstrate the effectiveness of the proposed approach through experiments on real robotic systems. 
\end{itemize}

\section{Related work}

The motion-planning problem is typically addressed by discretizing the continuous state space, using grid-based methods for graph searches or random sampling for stochastic searches. While graph-based algorithms like Dijkstra~\cite{dijkstra2022note} and A*\cite{duchovn2014path} are resolution-complete, they become computationally prohibitive in high-dimensional scenarios. Sampling-based planning alleviates this issue by exploring large areas efficiently without the high computational cost of traditional methods. Early sampling-based algorithms like PRM \cite{kavraki1996probabilistic}, EST \cite{hsu2002randomized}, and RRT \cite{lavalle1998rapidly} lack solution path asymptotically optimality. RRT*, introduced by Karaman and Frazzoli \cite{karaman2011sampling}, advanced optimal path planning by achieving asymptotic optimality. However, efficiently finding paths in high-dimensional spaces remains challenging. Recent approaches have enhanced RRT* with Edge Optimization and Sample Optimization to improve solution quality and convergence speed.

\subsection{Edge Optimization Strategies}

Edge optimization strategies incrementally improve path quality by refining connections within tree or graph structures, enabling efficient path planning in complex environments. Building on RRT*, RRT-sharp (RRT$^\#$)~\cite{arslan2013use} uses a more efficient rewiring cascade to propagate reduced cost-to-goal information within local neighborhoods, albeit with higher computational costs. Real-Time RRT* (RT-RRT*)\cite{naderi2015rt} supports rapid adjustments in dynamic environments without tree reconstruction, while RRTFN~\cite{tong2019rrt} balances tree reuse and new exploration for real-time efficiency. However, both methods struggle with path optimization tasks.
Recent edge optimization for high-dimensional spaces includes Pan et al.'s~\cite{pan2013faster} k-NN-based collision checking, enhancing sampling efficiency, and Kleinbort et al.'s~\cite{kleinbort2020collision} r-disc strategy, which can outperform k-NN in certain non-Euclidean, high-dimensional spaces. Though these algorithms yield locally optimal paths, nearest-neighbor constraints often limit global optimality. Quick-RRT*~\cite{jeong2019quick} improves global solutions through vertex ancestry, despite higher computational demands.
\textcolor{black}{This work introduces an adaptive ancestor strategy that selectively explores a subset of ancestors, reducing computational costs while preserving solution quality across dimensions. Furthermore, to efficiently identify low-cost trajectories, we propose optimizing the sampling strategy during point distribution to improve the efficiency performance further.}

\subsection{Sample Optimization Strategies}

Sample optimization strategies improve planning efficiency and path quality by refining sample generation and selection, minimizing redundant computations, and accelerating optimal path discovery.
For example, Intelligent Sampling~\cite{islam2012rrt} generates nodes near obstacle vertices, inspired by visibility graph techniques, optimizing node distribution and improving path planning in complex environments. Informed Sampling methods, like Informed RRT*\cite{gammell2014informed}, concentrate sampling within an ellipsoid around the current best path (informed set), accelerating convergence by focusing on promising areas. BIT*~\cite{gammell2015batch} enhances sample information growth by using batch sampling and initializing a random geometric graph (RGG). Its advanced version, ABIT*~\cite{strub2020advanced}, introduces inflation and truncation factors to balance exploration and exploitation within a dense RGG. 
\textcolor{black}{Recent state-of-the-art informed set methods like GuILD~\cite{scalise2023guild} and FIT*~\cite{zhang2024flexible} improve sampling efficiency by refining subset selection and adapting batch sizes. However, GuILD’s beacon-based selection may cause premature convergence in high-dimensional spaces, while FIT* scales batch sizes dynamically but is limited by predefined informed sets.}
\textcolor{black}{While EIT~\cite{strub2022adaptively} improves path optimization with heuristic functions, it discards entire batches upon collision detection, leading to inefficiencies. Our method addresses this by dynamically refining connectivity through the Just-in-Time Edge module, enhancing path quality, and reducing overall path length.
Recent methods like Elliptical K-Nearest Neighbors (FDIT*)~\cite{zhangfdit} and Fully Connected Informed Trees (FCIT*)~\cite{Wilson2024fcit} use nearest-neighbor techniques to enhance sampling-based planning. However, FDIT* still suffers from k-NN's inherent limitations, and despite FCIT* leveraging brute-force parallelism to bypass explicit neighbor searches, it remains constrained by nearest-neighbor strategies, limiting its adaptability.}
To address this, we propose a "Just-in-Time" sampling method: rather than discarding all points after a collision, we increase sampling density in frequently colliding areas. JIT* enhances feasible path discovery by adapting sampling density to obstacle interactions, improving navigation efficiency in complex environments.

\subsection{Sampling-based planning with kinematic constraint}

Kinematic constraints are essential for safely deploying sampling-based algorithms in robotic arms, ensuring efficient path planning in complex environments. Liu et al.\cite{liu2020improved} enhanced tree-based planning by integrating humanoid arm kinematics, while Liang et al.\cite{liang2020pr} introduced PR-RRT* for a 6-DOF arm, combining forward kinematics with probabilistic root-based sampling. Gao et al.~\cite{gao2023path} proposed a BP neural network-enhanced RRT* to optimize node sampling and collision avoidance. Though these methods emphasize kinematic safety, they often neglect manipulability, leading to singularities.
Shen~\cite{shen2023adaptive} incorporated manipulability into RRT* to avoid singularities, while Kaden~\cite{kaden2019maximizing} optimized paths for both manipulability and obstacle avoidance using an extended RRT with state costs. Zhang~\cite{zhang2024path} improved RRT*-based planning by integrating manipulability as prior information, enhancing collision avoidance and motion quality. However, these approaches do not account for multi-arm self-collisions, limiting their applicability. Additionally, relying on traditional RRT* without heuristic-guided sampling slows convergence in cluttered environments with narrow passages.





\section{Problem Formulation}

Feasible path planning identifies a sequence of valid states that successfully navigates from start to goal. While there are often multiple solutions, the proposed method targets the optimal one---finding a path that maximizes efficiency and motion performance. The optimal path planning problem is formally defined as:

\noindent
Given an initial and desired goal state, i.e.,  
$
\textcolor{black}{\mathbf{x}_{\text{start}}, 
\mathbf{x}_{\text{goal}} }
\in X_{free}$, 
where 
$X_{free} $ defines the free state space, satisfying 
$X_{free}:=\rm{closure}$($X$/$X_{obs}$)---with 
$ X_{obs}  $ 
being the obstacle space and 
$ X {\subseteq} \mathbb{R}^n $ the planning state space with  $n {\in} \mathbb{N}$---%
the objective of our path planning problem is to find  an optimal path $\pi^*$ within the set $\Pi$ of all valid paths $\pi: [0,1] \rightarrow X_{free}$ 
that satisfies
\begin{eqnarray}
\begin{array}{lllllll}
\pi^* = & \mathop {{\rm{argmin}}}\limits_{\pi  \in \Pi } \mathcal{S}_{\text{total}}(\pi ) \\

\textrm{s.t.}
&\pi(0) = \textcolor{black}{\mathbf{x}_{\text{start}}}, \pi(1) = \textcolor{black}{\mathbf{x}_{\text{goal}}}, \\ 
&\forall t \in [0, 1], \pi(t) \in \mathit{X}_{{free}}.&
\end{array}
\label{eq: objective function}
\end{eqnarray}
%



More specifically, the valid path $\pi$ consists of a sequence of edges, where each edge connects a source to a target state, respectively, $\mathbf{x_s}, \mathbf{x_t}$.  The total cost
$\mathcal{S}_{\text{total}}: \Pi {\rightarrow} \mathbb{R}_{\geq 0}$ 
is thus the sum of the dynamic heuristic costs $\mathcal{S}(\mathbf{x_s}, \mathbf{x_t})$ for each edge, i.e., 
%
%
%
\begin{equation} 
\label{reverse_search_cost}
\mathcal{S}_{\mathcal{R}}^{\mathrm{JIT}^*}(\mathbf{x_s}, \mathbf{x_t}) : \left\{
\begin{array}{l}
s^1_{sub}: \alpha C_{\text{sum}} + (1- \alpha) D_{\text{tanh}} \\
s^2_{sub}: \bar{e}(\textcolor{black}{\mathbf{x}_{\text{goal}}}, \mathbf{x_s}) + \bar{e}(\mathbf{x_s}, \mathbf{x_t}) + \bar{d}(\mathbf{x_t},\textcolor{black}{\mathbf{x}_{\text{start}}})
\end{array}
\right.
\end{equation}
\begin{equation}
C_{\text{sum}} = \hat{h}(\mathbf{x_s}) + \hat{c}(\mathbf{x_s}, \mathbf{x_t}) + \hat{g}(\mathbf{x_t})
\end{equation}
where cumulative cost $C_{\text{sum}}$ proposed for path efficiency including admissible cost-to-goal heuristic  $\hat{h}(\cdot)$, an inadmissible cost heuristic $\hat{c}(\cdot, \cdot)$, and the admissible prior cost-to-start $\hat{g}(\cdot)$, where $\hat{g}(\mathbf{x}):= \hat{c}(\textcolor{black}{\mathbf{x_{\text{start}}}}, \mathbf{x})$. A motion performance term, $D_{\text{tanh}}$, optimizes manipulability and safeguards against approaching singularities.
\textcolor{black}{The parameter $\alpha$ serves as a scaling factor that ensures consistent units. }$\bar{e}(\cdot,\cdot)$ measures the computational effort to validate paths between states, while $\bar{d}(\cdot)$ represents the inadmissible prior effort to start heuristics, defined as $\bar{d}(\mathbf{x}):= \bar{e}(\textcolor{black}{\mathbf{x_{\text{start}}}}, \mathbf{x})$. The term $s^1_{sub}$ serves as the highest-priority optimization objective, ensuring the cost heuristic remains admissible. Subsequent objectives $s^2_{sub}$ ensure tie-breaking favors paths with lower estimated effort. When the algorithm identifies an initial feasible path, $s^1_{sub}$ remains the primary optimization target in the $\mathcal{S}_{\mathcal{F}}^{\mathrm{JIT}^*}$ forward search. The proposed JIT* algorithm seeks to generate an optimal path by balancing path efficiency with motion performance.



\section{Methodology}

\begin{figure*}[htp] 
\centering 
\includegraphics[width=0.98\textwidth]{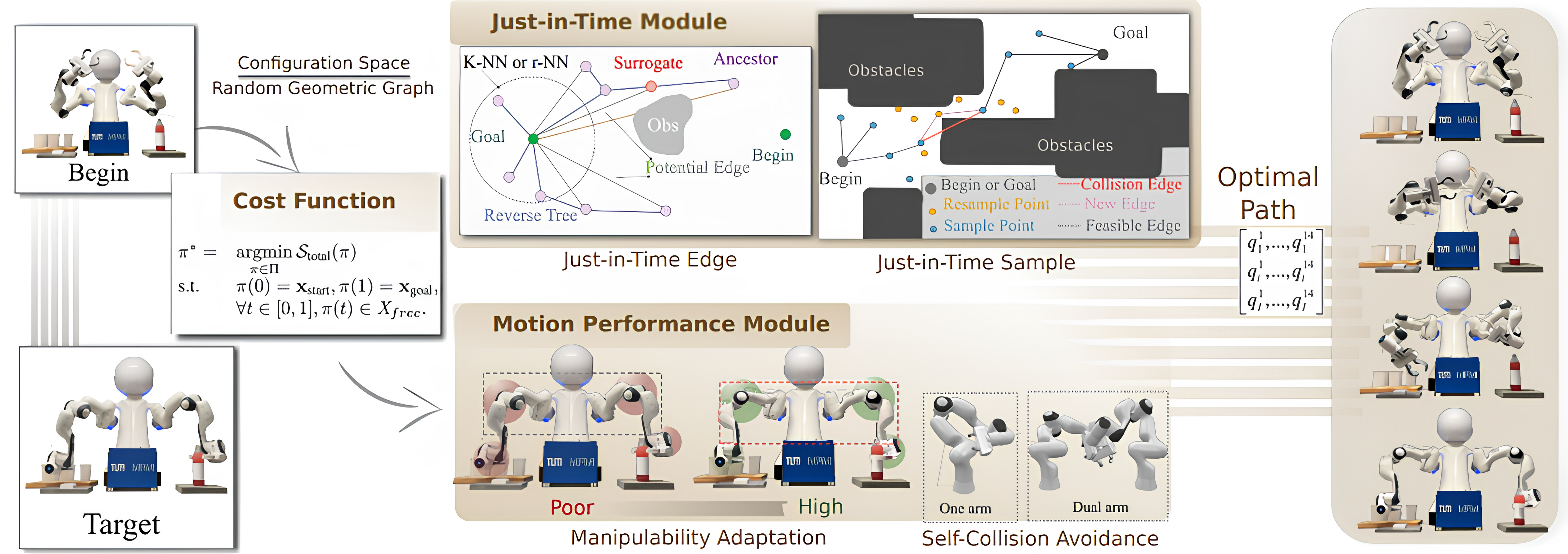}
\caption{System diagram of the proposed path planning method. After setting the task's start and end points, our algorithm optimizes the dynamic heuristic costs across each edge. It comprises two main modules: the Just-in-Time module, which enhances trajectory efficiency through Edge and Sample components, and the Motion Performance module, which optimizes manipulability and prevents self-collisions, ensuring the trajectory aligns with the manipulator’s dynamic constraints.}
\label{fram} 
\end{figure*}

\subsection{Framework of JIT*}

\begin{algorithm}[t]
\caption{Just-in-Time Informed Tree (JIT*)}
\label{Alg:frame}
\SetKwInOut{Input}{Input}
\SetKwInOut{Output}{Output}
\SetKwFunction{biasSample}{biasSample}
\SetKwFunction{bestKey}{bestKey}
\SetKwFunction{reverseSearch}{reverseSearch}
\SetKwFunction{restartReverseSearch}{restartReverseSearch}
\SetKwFunction{forwardSearch}{forwardSearch}
\SetKwFunction{couldImproveSolution}{couldImproveSolution}
\SetKwFunction{pathFound}{pathFound}
\SetKwFunction{lazyCheck}{lazyCheck}
\SetKwFunction{terminateCondition}{terminateCondition}
\SetKwFunction{justEdge}{justEdge}
\SetKwFunction{saveSaftyDomain}{saveSaftyDomain}
\SetKwFunction{prune}{prune}
\SetKwFunction{justSample}{justSample}
\SetKwFunction{isLeer}{isLeer}
\SetKwFunction{costObstacleLazyCheck}{costObstacleLazyCheck}
\SetKwFunction{isRGGDensitySufficient}{isRGGDensitySufficient}
\SetKwFunction{updateReverse}{updateReverse}
\SetKwFunction{fullCheck}{fullCheck}
\SetKwFunction{expandWithNeigborAndPRM}{expandWithNeigborAndPRM}
\SetKwFunction{updateForward}{updateForward}
\SetKwFunction{expendNeighbors}{expendNeighbors}
\SetKwFunction{addEdgeTarget}{addEdgeTarget}
\SetKwFunction{justEdge}{justEdge}
\SetKwFunction{LazyCheck}{LazyCheck}
\SetKwFunction{expandWithRGG}{expandWithRGG}
\SetKwFunction{setRGGDensitySufficient}{setRGGDensitySufficient}

\DontPrintSemicolon

\Input{$\text{Start point}~\textcolor{black}{\mathbf{x}_{\text{start}}}$, \text{goal point}~$\textcolor{black}{\mathbf{x}^*_{\text{goal}}}$, \textcolor{purple}{\text{DH parameters}}}
\Output{$\text{Optimal path}~\mathcal{\pi^*}$}
$X_{\textit{free}} \gets \{\textcolor{black}{\mathbf{x}_{\text{start}},\mathbf{x}^*_{\text{goal}}}\},$\\
$\mathcal{Q_F}\gets \expendNeighbors(\mathbf{x}_{\text{start}}),$

$\mathcal{Q_R} \gets \expendNeighbors(\textcolor{black}{\mathbf{x}^*_{\text{goal}}}),$\\
$E = (\mathbf{x}_{\text{s}},\mathbf{x}_{\text{t}}), \mathcal{T_F} = (\mathcal{V_F}),\mathcal{T_R} = (\mathcal{V_R}), $$\pi \gets \emptyset$\\
$\textcolor{purple}{E_{\mathcal{R}} \gets \underset{\mathcal{Q_R}}{\arg\min} \, S_{\mathcal{R}}^{\mathrm{JIT}^*},}$
$\textcolor{purple}{E_{\mathcal{F}} \gets \underset{\mathcal{Q_F}}{\arg\min} \, S_{\mathcal{F}}^{\mathrm{JIT}^*}}$\\

$X_{\textit{free}}, \textcolor{purple}{X_{\textit{obs}} \stackrel{+}{\leftarrow}} \textcolor{purple}{\biasSample()}$\\
\While{\textbf{not} $\terminateCondition()$}{
        \If{$\textcolor{black}{\LazyCheck(E_{\mathcal{R}})}$}
        {
            $\textcolor{purple}{\mathcal{Q_R}\gets \justEdge(E_{\mathcal{R}},\mathcal{T_R})}$\Comment{ancestor neighbors of reverse tree}\\
            $\textcolor{purple}{E_{\mathcal{R}} \gets \underset{\mathcal{Q_R}}{\arg\min} \,\mathcal{S}_{\mathcal{R}}^{\mathrm{JIT}^*}}$\Comment{sort reverse edge via heuristics}\\
            $\mathcal{T_R} \gets \updateReverse(E_{\mathcal{R}})$\\           
            \If{$\couldImproveSolution(\mathcal{T_R})$}{
                \eIf{$\fullCheck(E_{\mathcal{F}})$}{
                        $\textcolor{purple}{\mathcal{Q_F} \gets \expandWithRGG()}$\Comment{expand neighbors of forward tree}\\ $\mathcal{T_F} \gets \updateForward(E_{\mathcal{F}})$\\
                        $\textcolor{purple}{E_{\mathcal{F}} \gets \underset{\mathcal{Q_F}}{\arg\min} \, \mathcal{S}_{\mathcal{F}}^{\mathrm{JIT}^*}}$\Comment{sort forward edge via heuristics}\\
                        $\textcolor{purple}{\mathcal{Q_F}\gets \justEdge(E_{\mathcal{F}},\mathcal{T_F})}$\Comment{update ancestor neighbors of forward tree}\\

                    }{
                        \textcolor{purple}{$X_{\textit{free}}, X_{\textit{obs}} \stackrel{+}{\leftarrow} \justSample()$}\Comment{add samples into the c-space}\\
                        $\restartReverseSearch()$\\
                        
                    }
                }
            
        }
$ \pi^* \leftarrow \prune(\mathcal{X}_{\textit{free}}, \textcolor{purple}{X_{\textit{obs}}}, \mathcal{T_F})$

$X_{\textit{free}}, \textcolor{purple}{X_{\textit{obs}} \stackrel{+}{\leftarrow}} \textcolor{purple}{\biasSample()}$\\

        }{\Return {$Optimal Path \ \pi^*$}

}
\end{algorithm}

The JIT* algorithm leverages a random geometric graph (RGG) generated through a lazy tree in reverse search, $\mathcal{T_R} = (V_\mathcal{R}, E_\mathcal{R})$, to inform and guide the incremental tree in the forward search, $\mathcal{T_F} = (V_\mathcal{F}, E_\mathcal{F})$, toward finding a path.  Based on this initial path, JIT* iteratively restarts the reverse and forward search to further optimize until it identifies the optimal path $\pi^*$. The vertices in these trees, $V_\mathcal{F}$ and $V_\mathcal{R}$, correspond to free states in $X_{free}$. Edges in the forward tree, $E_\mathcal{F} \subseteq V_\mathcal{F} \times V_\mathcal{F}$, represent valid state connections, while edges in the reverse tree, $E_\mathcal{R} \subseteq V_\mathcal{R} \times V_\mathcal{R}$, may traverse invalid regions of the problem domain. 
Each edge consists of a source state, $\mathbf{x}_s$, and a target state, $\mathbf{x}_t$, denoted as $(\mathbf{x}_s, \mathbf{x}_t)$. 

JIT* initializes the start state $\textcolor{black}{\textbf{x}_{\text{start}}} \in X_\text{free}$, the updated goal state $\textcolor{black}{\textbf{x}^*_{\text{goal}}}\in X_\text{free}$, and the reverse and forward queues, $\mathcal{Q_R}$ and $\mathcal{Q_F}$. Here, $\textcolor{black}{\textbf{x}^*_{\text{goal}}}$ represents a goal optimized for manipulability, maintaining the end-effector's position while enhancing the manipulator's configuration. Further details are provided in Section~\ref{MI}. If a direct path between $\textcolor{black}{\textbf{x}_{\text{start}}}$ and $\textcolor{black}{\textbf{x}^*_{\text{goal}}}$ is infeasible, the function \texttt{biasSample()} is designed to sample within the current batch size, taking self-collision into account to explore feasible areas effectively.
The notation $\stackrel{+}{\leftarrow}$ in Line 6 of Alg.~\ref{Alg:frame} is used to incorporate sampled points into $X_\text{free}$ and $X_\text{obs}$. Following sampling, the reverse search begins with the cost function $\mathcal{S}_{\mathcal{R}}^{\mathrm{JIT}^*}$, improving the initial solution (Lines 9-11 in Alg.~\ref{Alg:frame}).
In the reverse tree, \texttt{LazyCheck()} evaluates edge cost and motion performance based on environmental properties. 
To overcome typical neighbor connection limits and reduce path length, \texttt{justEdge()} processes selected edges to extend promising connections. \texttt{updateReverse()} reconstructs branches in $\mathcal{T_R}$, and the search continues by balancing efficiency and motion performance until a path from $\textcolor{black}{\textbf{x}^*_{\text{goal}}}$ to $\textcolor{black}{\textbf{x}_{\text{start}}}$ is identified.
After the reverse path is established (Alg.~\ref{Alg:frame}, Lines 12–19), the forward search begins with \texttt{couldImproveSolution()}, checking if new edges reduce path length. It then calls \texttt{expandWithRGG()} to balance cost and motion performance. Upon finding a solution, the search pauses for path refinement. If no solution emerges due to collisions, \texttt{justSample()} targets bottleneck regions to overcome critical constraints.
\textcolor{black}{JIT* terminates when the \texttt{terminateCondition}—either a maximum time limit for trajectory planning and optimization or a maximum iteration count—is met, yielding the optimal path, as illustrated in Fig.~\ref{fram}.}



\subsection{Just-in-Time module}
\label{JIT_Framework}

Robotic environments are often complex, challenging traditional sample-based planning methods that rely on uniform or goal-biased sampling. To address these challenges, we introduce a Just-in-Time model within a bidirectional lazy search tree framework~\cite{strub2022adaptively}. \textcolor{black}{Lazy reverse search provides global guidance with low-cost heuristics, while the forward tree ensures feasibility under constraints. This approach minimizes unnecessary collision checks, significantly improving efficiency by reducing redundant edge validations, a major computational bottleneck in sampling-based motion planning.} Based on this, our method enhances pathfinding efficiency through two strategies: "Just-in-Time edge" and "Just-in-Time sample," optimizing connectivity and sampling in critical areas. Unlike EIT*, which inefficiently discards entire sample batches upon collision, JIT adaptively densifies sampling in bottlenecks and refines edge connections for faster convergence to optimal paths.

\subsubsection{Just-in-Time Edge Expend}
\label{JIT_Framework}

The Just-in-Time edge expansion dynamically establishes connections between a node and selectively chosen ancestors based on an 
\textbf{adaptive ancestor strategy} to expand the search space.  
Rather than traversing all ancestors---
which is computationally intensive---it 
selects a subset of ancestor nodes guided by collision detection and bounded by a predefined threshold. This reduces computational overhead in reverse search while preserving solution quality.
This reduces reverse search overhead while maintaining solution quality, overcoming the rigidity of fixed $k$-NN and $r$-NN approaches by adapting the search space (Fig.~\ref{fram}).

\noindent{\textbf{Definition 1.} (Ancestor set)}\
\textit{For a vertex $\mathbf{x}$ in a tree $\mathcal{T}$, the $k$-th ancestor, $\text{a}_k(\mathbf{x})$, is the node reached by moving $k$ levels up from $\mathbf{x}$ toward the root. The parent of $\mathbf{x}$ is given by $\text{a}_0(\mathbf{x}) = \mathrm{parent}(\mathbf{x})$. Formally, the recursive relationship is defined as:
\begin{equation} 
\text{a}_k(\mathbf{x}) = \mathrm{parent}(\text{a}_{k-1}(\mathbf{x})), \quad \text{for all } k > 0.
\end{equation}}

\begin{algorithm}[t]
\caption{JIT*:\ Just-in-Time Edge}
\label{Alg: adaptiveNeighbors}
\DontPrintSemicolon 
\SetKwFunction{findNeighbors}{findNeighbors}
\SetKwFunction{parent}{parent}
\SetKwFunction{children}{children}
\SetKwFunction{exist}{exist}
\SetKwFunction{collisionFree}{collisionFree}
\SetKwFunction{cost}{cost}
\SetKwFunction{Sample}{Sample}
\SetKwFunction{add}{add}
\SetKwFunction{getEdge}{getEdge}

$V_{\mathrm{neighbors}} \gets \findNeighbors(\mathbf{x},\ r\ \text{or}\ k)$\;
$V_{\mathrm{ancestors}} \gets \emptyset$\;
$c_{\mathrm{step}} \gets 0$\;
$\mathbf{x}_{\mathrm{tmp}} \gets \parent(\mathbf{x})$\;
$\mathbf{x}_{\mathrm{prev}} \gets \mathbf{x}$\;

\While{$\mathbf{x}_{\mathrm{tmp}} \neq \emptyset$ \ \textbf{and} \ $c_{\mathrm{step}} \leq \tau$}{
    $c_{\mathrm{step}} \gets c_{\mathrm{step}} + 1$\Comment{iteration counts of the loop}\;
    
    \eIf{\collisionFree{$\mathbf{x}$, $\mathbf{x}_{\mathrm{tmp}}$}}{
        $V_{\mathrm{ancestors}} \stackrel{+}{\leftarrow} \{\mathbf{x}_{\mathrm{tmp}}\}$\Comment{insert temp node into ancestor}\;
        $\mathbf{x}_{\mathrm{prev}} \gets \mathbf{x}_{\mathrm{tmp}}$\Comment{update previous valid node}\;
        $\mathbf{x}_{\mathrm{tmp}} \gets \parent(\mathbf{x}_{\mathrm{tmp}})$\Comment{update parent node of the current temp node}\;
    }
    {
        $X_{\mathrm{se}} \gets \Sample(\getEdge(\mathbf{x}_{\mathrm{prev}}, \mathbf{x}_{\mathrm{tmp}}))$\Comment{sample new node into space}\;
        \ForEach{$\mathbf{x}_{\mathrm{samp}}$ \textbf{in} $X_{\mathrm{se}}$}{
            \If{\collisionFree{$\mathbf{x}$, $\mathbf{x}_{\mathrm{samp}}$}}{
                $V_{\mathrm{ancestors}} \stackrel{+}{\leftarrow} \{\mathbf{x}_{\mathrm{samp}}\}$\Comment{insert new valid sample node}\;
                \textbf{break}\;
            }
        }
        \textbf{break}\;
    }
}
$V_{\mathrm{neighbors}} \stackrel{+}{\leftarrow} V_{\mathrm{ancestors}}$\;
$\mathcal{Q}_{\mathrm{R}} \stackrel{+}{\leftarrow} V_{\mathrm{neighbors}}$\;

\Return{$\mathcal{Q_R}$}\;
\end{algorithm}

\textcolor{black}{Instead of including all ancestors, we iteratively evaluate each ancestor $\text{a}_k(\mathbf{x})$ by checking for a direct, collision-free connection from vertex $\mathbf{x}$. If such a connection exists, the ancestor is added to the set $V_{\text{ancestors}}$ (see Alg.~\ref{Alg: adaptiveNeighbors}, lines 8–11). This process continues until the cumulative step count $c_{\text{step}}$ exceeds the threshold $\tau$ or no more ancestors are available, ensuring bounded computational complexity.
If a collision is detected during a connection attempt, the algorithm samples intermediate points along the edge between the last valid ancestor $\mathbf{x}_{\text{prev}}$ and the current ancestor $\mathbf{x}_{\text{tmp}}$ (line 13). For each sampled point $\mathbf{x}_{\text{samp}}$, a collision-free connection to vertex $\mathbf{x}$ is attempted. If successful, the sampled point is added to the ancestor set as a surrogate for the colliding ancestor, and the process stops.
This approach balances path length and computational cost by selectively expanding the ancestor set while minimizing redundant evaluations for other ancestors once a collision is detected.}

\subsubsection{Just-in-Time Sample Strategy}
\label{JIT_Framework}


Obstacle-biased sampling is commonly used in path planning to improve success rates in dense obstacle environments. However, not all obstacles require attention, and this approach can lead to unnecessary detours. In this paper, the proposed method integrates with bidirectional lazy tree and batch sampling frameworks. Focusing on critical sampling regions avoids wasted effort in non-critical obstacle areas. These priority regions occur where reverse search succeeds but forward search fails, typically due to narrow passages acting as bottlenecks or a lack of viable paths. Our approach increases sample density in these regions, guiding reverse search to explore them more thoroughly, as illustrated in Fig.~\ref{fram}.

\noindent \textbf{Definition 2.} (Priority sampling region)\, \textit{The priority sampling region around a potential edge is defined as the intersection of state spaces surrounding the edge's source and target, with a range determined by cost. This region, $X_{\hat{R}}$, lies within the space of all possible paths that can improve the current best solution, known as the informed set $X_{\hat{f}}$~\cite{gammell2018informed}.}
\begin{equation}
X_{\hat{R}} = B^n_{{s, c}} \cap B^n_{{t, c}} \cap X_{\hat{f}},
\end{equation}
\begin{equation}
X_{\hat{R}} := \left\{ \mathbf{x} \in X_{\text{free}} : \|\mathbf{x} - \mathbf{x_s}\|_2, \|\mathbf{x} - \mathbf{x_t}\|_2 < c(\mathbf{x_s}, \mathbf{x_t}) \right\},
\end{equation}
where $B^n_{{s, c}}$ and $B^n_{{t, c}}$ are hyperspheroid centered at \(\mathbf{x_s}\) and \(\mathbf{x_t}\) respectively, and this region can be divided to two hyperspherical caps in $n$ dimension. The formula for the Lebesgue measure $\lambda(\cdot)$ of $X_{\hat{R}}$ is expressed as follows:
\begin{align}
\lambda(X_{\hat{R}}) = & \ 2\zeta_n r^n \left( \frac{1}{2} - \frac{r - h}{r} \frac{\Gamma\left(1 + \frac{n}{2}\right)}{\sqrt{\pi} \Gamma\left(\frac{n+1}{2}\right)} \right. \notag \\
& \left. {}_2F_1\left( \frac{1}{2}, 1 - \frac{n}{2}; \frac{3}{2}; \left(\frac{r-h}{r}\right)^2 \right) \right),
\end{align}
where the unit \( n \)-ball is \(\zeta_n\) and hypergeometric function is \( {}_2F_1 \). The height $h=\frac{1}{2}c(\mathbf{x}_\text{s}, \mathbf{x}_\text{t})$ and radius $r=c(\mathbf{x}_\text{s}, \mathbf{x}_\text{t})$. The function \(\zeta_n\) is given by:
\begin{equation}
\zeta_n = \frac{\pi^{\frac{n}{2}}}{\Gamma\left(\frac{n}{2} + 1\right)},
\end{equation}
where \(\Gamma(z)\)is the gamma function, an extension of factorials to real numbers.





\subsection{Motion performance module}

\subsubsection{Manipulability improvement}
\label{MI}

Manipulability is a key concept in the motion performance module, capturing the robot's capacity to maneuver its end-effector with precision and joint-to-task space efficiency throughout its workspace \cite{1985_Yoshikawa_IJRR,1997_Chiaverini_TRA,Kurtz2021}. It reflects the joint effort required to  
move or apply forces in different directions, making it essential for task feasibility, performance, and control stability. 
While different metrics have been explored for performance evaluation, see, e.g.,  \cite{Patel2014},  the manipulability functions based on the determinant, i.e.,  
 $\sqrt{\text{det} J J^{T}}$, and the min. singular value ($\sigma_{\text{min}}(J)$) of the robot's geometric Jacobian are the most widely employed. We favor the latter as the determinant lacks monotonicity, as it gets closer to singularities, and the  $\sigma_{\text{min}}$ optimizes the performance considering the worst-case scenario. This provides a more stable motion.   
In our design, it is integrated in (Eq.\ref{reverse_search_cost}) through the heuristic,  %
%
\begin{equation}
D_{\text{tanh}} = \tanh(\eta / (\sigma_{\text{min}}+ \varepsilon )) / (\sigma_{\text{min}}+ \varepsilon ),
\end{equation}
where $\eta$ is the constant value, 
$\varepsilon {\geq} 0$ is a small constant to avoid division by zero, and  
$\sigma_{\text{min}}$ represents the minimum singular value 
\begin{equation}
\sigma_{\min}(\mathbf{x}) = \min_{ \mathbf{v} \in \mathbb{R}^n, \|\mathbf{v}\| = 1} \| J(\mathbf{x}) \mathbf{v} \|,
\end{equation}
where  $\|\cdot\|$ denotes the Euclidean norm. This formula is essentially aimed at identifying a unit vector $\mathbf{v}$ such that the Jacobian matrix $J(\mathbf{x})$ exerts minimal influence on $\mathbf{v}$. In other words, the objective is to find the direction $\mathbf{v}$ for which the magnitude of the resultant vector $J(\mathbf{x}) \mathbf{v}$ is minimal---reflecting poor motion capability in such direction. This magnitude, $\| J(\mathbf{x}) \mathbf{v} \|$, corresponds precisely to the minimum singular value, denoted as $\sigma_{\min}(\mathbf{x})$. Therefore, $\sigma_{\min}(\mathbf{x})$ can be calculated by evaluating the eigenvalues $\lambda_i(\mathbf{x})$ of the Jacobian matrix $J(\mathbf{x}) \in \mathbb{R}^{m \times n}$, which maps joint velocities to the cartesian velocities of the end-effector. By applying Singular Value Decomposition (SVD) to $J(\mathbf{x})$, we obtain $\sigma_{\min}(\mathbf{x})$, 
\begin{align}
J(\mathbf{x}) & = U(\mathbf{x}) \Sigma(\mathbf{x}) V^T(\mathbf{x}), 
\nonumber 
\\
\Sigma & = \text{diag}(\sigma_1, \sigma_2, \dots, \sigma_{\min(m,n)}).
\end{align}
where \( U(\mathbf{x}) \in \mathbb{R}^{m \times m} \) and \( V(\mathbf{x}) \in \mathbb{R}^{n \times n} \) are orthogonal matrices, and \( \Sigma(\mathbf{x}) \in \mathbb{R}^{m \times n} \) is a diagonal matrix containing the singular values \( \sigma_i(\mathbf{x}) \). $\sigma_{\min}(\mathbf{x})$ is the last element of the diagonal from $\Sigma$.These singular values represent the semi-axes of the manipulability ellipsoid, providing a direct measure of the robot’s capacity to execute movements in specific directions.

\textcolor{black}{The \( D_{\text{tanh}} \) function adapts to minimum singular values, guiding robotic path planning away from potential singularities for improved efficiency and safety. By incorporating a hyperbolic tangent function, \( D_{\text{tanh}} \) provides smooth scaling, adjusting responsively to variations in \( \sigma_{\text{min}} \).
When selecting the lowest-cost state for connection using \( s^1_{\text{sub}} \) in \eqref{reverse_search_cost},  the parameter \(\eta\) controls the emphasis on manipulability. Specifically, as the state's minimum singular value nears singularity, \( D_{\text{tanh}} \) outweighs \( C_{\text{sum}} \), prioritizing manipulability. Conversely, path length takes precedence as the value moves away from singularity.
This approach enables JIT* to balance cost with the robot’s kinematic constraints, ensuring paths that are both cost-effective and manipulability-efficient.}

In addition, our method includes a post-processing step that utilizes the Jacobian matrix’s null space. This allows the robot to maintain the end-effector position while adjusting joint configurations to improve manipulability. By leveraging null-space adjustments, the robotic system can transition to configurations that enhance overall motion performance without affecting the primary task. To ensure solution stability and manipulability, we check if the minimum singular value \( \sigma_{\text{min}} \) of the Jacobian matrix $J(\mathbf{x_{goal}})$ falls below a predefined threshold \( \epsilon \). This is done using the indicator function defined as:
\begin{equation}
\mathbb{M}(\sigma_{\text{min}(J(\mathbf{x_{goal}}))}) = \mathbb{I}(\sigma_{min} > \epsilon),
\end{equation}
where \( \mathbb{I}(\cdot) \) is the indicator function. If the indicator function returns 0, meaning \( \sigma_{\text{min}} \leq \epsilon \), it indicates that the goal is at or near a singularity. In such cases, the dimension of the null space of the Jacobian matrix $d$ can be calculated as:
\begin{equation}
d = n - \sum_{i=1}^{\min(m,n)}\mathbf{I}(\sigma_i > 0).
\end{equation}
The basis of the null space is composed of the last $d$ rows of the matrix $V(x)$, which forms the null space basis matrix $\mathbf{N}$. After that, we defined a perturbation vector $\mathbf{\lambda}_i$ to explore arm configurations within the null space, which aims to identify a new configuration that maximizes manipulability while ensuring that the end-effector's position remains unchanged. The formula of the updated configuration $\mathbf{x^*_{goal}}$ is defined as:
\begin{equation}
\textcolor{black}{\mathbf{x^*_{\text{goal}}}} = \arg\max_{i=1,\dots,I} \sigma_{\min}(J(\textcolor{black}{\mathbf{x_{\text{goal}}}} + \mathbf{\Delta x}_i)).
\end{equation}
\begin{equation}
\mathbf{\Delta x}_i = \mathbf{N}^T \mathbf{\lambda}_i,
\end{equation}

In sample-based planning like Informed RRT*, interpolated trajectories between tree nodes must maintain high manipulability for stable execution. We optimize the manipulability of interpolated states $\mathbf{x}_{\text{inter}}=[x_1, x_2, ..., x_n]$ within the trajectory, adjusting the interpolation state in configuration space $\mathbf{x}^*_{\text{inter}}$ as:
\begin{equation}
\mathbf{x}^*_{\text{inter}} = \mathbf{x}_{\text{inter}} + \mathcal{N}(\mu, \tau^2)(I - J^{\dagger}(\mathbf{x}_{\text{inter}})J(\mathbf{x}_{\text{inter}})) \dot{\mathbf{x}}_{\sigma_{\text{min}}},
\end{equation}
\begin{equation}
\dot{\mathbf{x}}_{\sigma_{\text{min}}} = J^{\dagger}_{\sigma_{\text{min}}} \sigma_{\text{min}}(J),
\end{equation}
where \( J^{\dagger}_{\sigma_{\text{min}}} \) is the pseudo-inverse of the Jacobian matrix, adjusted to minimize the effect of the smallest singular value \( \sigma_{\text{min}}(J) \), enhancing stability. \(\mathcal{N}\) is a Gaussian model with mean $\mu=0.5$ and covariance $\tau=0.4$, determined through offline tests.  
This method maximizes manipulability within the null space, ensuring stability and avoiding dynamically unstable configurations. By minimizing low singular value effects, the robot maintains task compliance and robustness throughout execution. While we adopt null-space methods to enhance manipulability, they introduce minor computational.



\subsubsection{Refinement of Self-Collision Check Bias Sampling}
\label{self-collision}



Self-collision avoidance is critical for both single-arm and multi-arm robotic systems, as unexpected collisions can lead to task failure or damage. Ensuring arms operate without interference is essential for control reliability and safety.

We introduce Self-Collision Danger Fields (SCDF) into bias sampling, quantifying collision risks between links within a robotic arm or across multiple arms. Non-adjacent edges are represented as $\overline{P_1P_2}$ and $\overline{Q_1Q_2}$, with any point ($\mathbf{P}(u_1)$ and $\mathbf{Q}(u_2)$) parameterized by scalar values $u_1$ and $u_2$ as follows:
\begin{equation}
\mathbf{P}(u_1) = P_1 + u_1 (P_2 - P_1), \quad u_1 \in [0, 1],
\end{equation}
\begin{equation}
\mathbf{Q}(u_2) = Q_1 + u_2 (Q_2 - Q_1), \quad u_2 \in [0, 1],
\end{equation}
where $P_1$ ($P_2$) and $Q_1$ ($Q_2$) represent the beginning pose (end pose) on the links $\overline{P_1P_2}$ and $\overline{Q_1Q_2}$, respectively. \textcolor{black}{The distance between points $\mathbf{P}(u_1)$ and $\mathbf{Q}(u_2)$ on the respective segments is given by:
\begin{equation}
\small
\begin{aligned}
&d^\mathbf{P}_\mathbf{Q} = \underbrace{(\mathbf{P}_2 - \mathbf{P}_1) \cdot (\mathbf{P}_2 - \mathbf{P}_1)}_{\alpha} u_1^2 + \underbrace{(\mathbf{Q}_2 - \mathbf{Q}_1) \cdot (\mathbf{Q}_2 - \mathbf{Q}_1)}_{\beta} u_2^2 \\
&+ \underbrace{-2(\mathbf{P}_2 - \mathbf{P}_1) \cdot (\mathbf{Q}_2 - \mathbf{Q}_1)}_{\gamma} u_1u_2 + \underbrace{2(\mathbf{P}_1 - \mathbf{Q}_1) \cdot (\mathbf{P}_2 - \mathbf{P}_1)}_{\delta}u_1 \\
&+ \underbrace{-2(\mathbf{P}_1 - \mathbf{Q}_1) \cdot (\mathbf{Q}_2 - \mathbf{Q}_1)}_{\epsilon} u_2 + \underbrace{(\mathbf{P}_1 - \mathbf{Q}_1) \cdot (\mathbf{P}_1 - \mathbf{Q}_1)}_{\zeta}.
\end{aligned}
\normalsize
\label{eq:long}
\end{equation}}
\textcolor{black}{To identify the critical points where the distance function $d^\mathbf{P}_\mathbf{Q}(u_1, u_2)$ is minimized, we take the first-order partial derivatives with respect to the parameters $u_1$ and $u_2$:
\begin{equation}
\begin{aligned}
\frac{\partial d^\mathbf{P}_\mathbf{Q}(u_1, u_2)}{\partial u_1} & = 2\alpha u_1 + \gamma u_2 + \delta = 0, \\
\frac{\partial d^\mathbf{P}_\mathbf{Q}(u_1, u_2)}{\partial u_2} & = 2\beta u_2 + \gamma u_1 + \epsilon = 0.
\end{aligned}
\end{equation}}
\textcolor{black}{Thus, the critical values $u_1^*$ and $u_2^*$, representing the closest points on the links, are obtained by solving the following equation, where $\gamma$ and $\beta$ are defined in Eq.~\ref{eq:long}:
\begin{equation}
\begin{pmatrix}
2\alpha & \gamma \\
\gamma & 2\beta
\end{pmatrix}
\begin{pmatrix}
u_1^* \\
u_2^*
\end{pmatrix}
=
\begin{pmatrix}
-\delta \\
-\epsilon
\end{pmatrix}.
\end{equation}}
\noindent\textbf{Notice}: The critical points $u_1^*$ and $u_2^*$ must satisfy the inequalities $0 \leq u_1^* \leq 1$ and $0 \leq u_2^* \leq 1$. If either $u_1^*$ or $u_2^*$ falls outside this interval, the points lie beyond the physical boundaries of the robot links and cannot be used for valid shortest distance evaluation. 
Considering such a situation, the overall minimum distance squared, denoted as $d^{\mathbf{P}*}_\mathbf{Q}$, is determined by the following 
\begin{equation}
d^{\mathbf{P}*}_\mathbf{Q} = \min \left\{ d^2(0, 0), d^2(0, 1), d^2(1, 0), d^2(1, 1), d^2(u_1^*, u_2^*) \right\}. \nonumber 
\end{equation}

The SCDF function $\Phi(\mathbf{A, B})$ quantifies the self-collision risk between arms $\mathbf{A}$, with $I$ links, and $\mathbf{B}$, with $J$ links, by evaluating the proximity between individual links within a single robotic arm or across multiple robotic arms. $\Phi(\mathbf{A,B})$ is expressed as 
\begin{equation}
\Phi (\mathbf{A,B}) {=}
\begin{cases} 
\sum_{i=1}^{I} \sum_{j=1}^{J}\frac{\kappa }{\sqrt{d_i^{j*}}} & \text{Risk between arms (1)}, \\
\sum_{i=1}^{I-2} \frac{\kappa }{\sqrt{d_i^{(i+2)*}}} & \text{Risk between links (2)},
\end{cases}
\label{eq:calculate risk}
\end{equation}
where $\kappa$ is the positive constant. This field calculates the self-collision risk between arms and between links.
We integrate this field into sampling to enhance self-collision avoidance, as detailed in Alg.~\ref{alg:check_self_collision}. A key parameter is the collision tolerance $\lambda = 0.018$, ensuring a minimum clearance for safe path execution.


\begin{algorithm}[tp]
    \caption{JIT*: Bias Sampling for Self-collision}
    \label{alg:check_self_collision}
    \SetKwFor{For}{for}{do}{endfor}
    \SetKwIF{If}{ElseIf}{Else}{if}{then}{else if}{else}{endif}
    \SetKw{Return}{return}
    \SetKw{Break}{break}
    \SetKw{Continue}{continue}
    
    \While{SampleProcess == True}{
        $\textbf{x}_{\text{sample}} \gets \Call{SampleConfiguration}{}$\;
        
        $links(begin, end) \gets \Call{ForwardKinematics}{\textbf{x}_{\text{sample}}}$\;
        
        \For{each non-adjacent links}{
            SelfCollisionRisk = equation (\ref{eq:calculate risk}) (2)\;
            \If{$SelfollisionRisk \geq \lambda$}{
                \Break  
            }
        }
        
        \For{each arms}{
            SelfCollisionRisk = equation (\ref{eq:calculate risk}) (1)\;
            \If{$SelfCollisionRisk \geq \lambda$}{
                \Break  
            }
        }
        
        \If{$SelfCollisionRisk < \lambda$}{
            \Return \textbf{SampleProcess = False}  
        }
    }
\end{algorithm}

\section{Analysis}

\subsection{Probabilistic Completeness}

\textbf{Theorem 1:} \textit{If a solution exists for this path planning problem with infinite samples, the probability that JIT* finds a feasible solution approaches one:}
\begin{equation}
\liminf_{i \to \infty} \mathbb{P}(\pi_i \in \Pi, \pi_i(0) = \mathbf{x_\text{start}}, \pi_i(1) = \mathbf{x_\text{goal}}) = 1,
\end{equation}
\textit{where $i$ is the number of samples, and $ \pi_i$ is the path from the initial state to the goal state discovered by the algorithm using those samples.}

\noindent \textbf{Proof.} The probabilistic completeness of JIT* follows from the proof of almost-sure asymptotic optimality in informed sampling-based planners. As the number of samples $\pi_i$ approaches infinity, the sampling process incrementally covers the state space. Since the underlying planner is probabilistically complete, JIT* retains this property, and its region refinement preserves this property.





\subsection{Almost-sure asymptotic optimality}

\textbf{Theorem 2:} \textit{If an optimal solution exists for this path planning problem, the probability that JIT* asymptotically converges to this optimal solution with infinite samples approaches one:}
\begin{equation}
\limsup_{q \to \infty} \mathbb{P}(c(\sigma_{q,\text{JIT*}}) = c(\sigma^*)) = 1,
\end{equation}
\textit{where $q$ is the sample count, $\sigma_{q,\text{JIT*}}$ is the path found by JIT* with $q$ samples, and $\sigma^*$ is the optimal solution.}

JIT* follows the same choose-parent and rewire strategies, with the rewiring radius \( r(q) \) satisfying:
\begin{equation}
r(q) > \eta \left(2 \left(1 + \frac{1}{d}\right)\left(\frac{\text{max}(\lambda(X_{\hat{f}}),\lambda(X_{\hat{r}}))}{\zeta_d}\right) \left( \frac{\log(q)}{q}\right)\right)^{\frac{1}{d}},
\end{equation}
where $\eta > 1$ is a tuning parameter, \( \lambda(\cdot) \) denotes the Lebesgue measure, \( d \) is the workspace dimension, \( \lambda(X_{\hat{f}}) \) and \( \lambda(X_{\hat{r}}) \) are measures of the informed sets, and \( \zeta_d \) is the unit ball volume.
This formula’s asymptotic optimality is proven for informed tree star-based algorithms~\cite{gammell2018informed}. JIT* preserves efficient rewiring and neighbor connections, adding only promising edges to enhance search. It increases sample density without compromising exploration or causing local optima entrapment. For manipulability, JIT* dynamically adjusts edge priorities while ensuring thorough exploration of essential connections.

\section{Experiments}\label{sec:Expri}
\subsection{Experiment Setup}


This section presents the quantitative analysis and experimental evaluation of our framework. Planning algorithms were developed using Planner Developer Tools (PDT)~\cite{gammell2022planner} and MoveIt~\cite{gorner2019moveit} for benchmarking. JIT* was compared with state-of-the-art (SOTA) algorithms in simulations and real-world experiments, including RRT-Connect~\cite{kuffner2000rrt}, $\text{RRT}^\#$~\cite{ArslanRRTsharp}, Informed RRT*~\cite{gammell2014informed}, BIT*~\cite{gammell2015batch}, ABIT*~\cite{strub2020advanced}, AIT*~\cite{strub2020adaptively}, and EIT*~\cite{strub2022adaptively}. Implementations were based on the Open Motion Planning Library (OMPL)~\cite{sucan2012open} and the ROS Noetic release on Ubuntu 20.04 LTS, running on an Intel Core i7-8750HQ with 16GB RAM. Planning terminates upon reaching the maximum computation time, with success defined as the current position $\mathbf{x}_{c}$ being within 0.01 of the target.

\subsection{Evaluation metrics}

The experimental analysis consists of two parts. The first evaluates planner efficiency, as informed-star-based planners rapidly find an initial solution and refine it over time. To assess performance, we recorded the \textbf{success rate}, \textbf{initial solution time} ($t^\textit{med}_\textit{init}$), \textbf{initial solution cost} ($c^\textit{med}_\textit{init}$), and \textbf{final solution cost} ($c^\textit{med}_\textit{final}$). $t^\textit{med}_\textit{init}$ and $c^\textit{med}_\textit{init}$ capture the time and path length of the first feasible solution, while $c^\textit{med}_\textit{final}$ reflects the optimized path length within the computation limit. The \textbf{success rate} and \textbf{cost trends} over 100 trials were tracked to evaluate overall performance.
The second part examines \textbf{motion performance}, measured by the \textbf{minimum singular value} ($\sigma_{\text{min}}$), which indicates the weakest movement direction. A higher $\sigma_{\text{min}}$ ensures strong manipulability and dexterity across all directions.


\begin{figure}[t!]
    \centering
    \begin{tikzpicture}
    \node[inner sep=0pt] (russell) at (-4.0,0.0)
    {\includegraphics[width=0.24\textwidth]{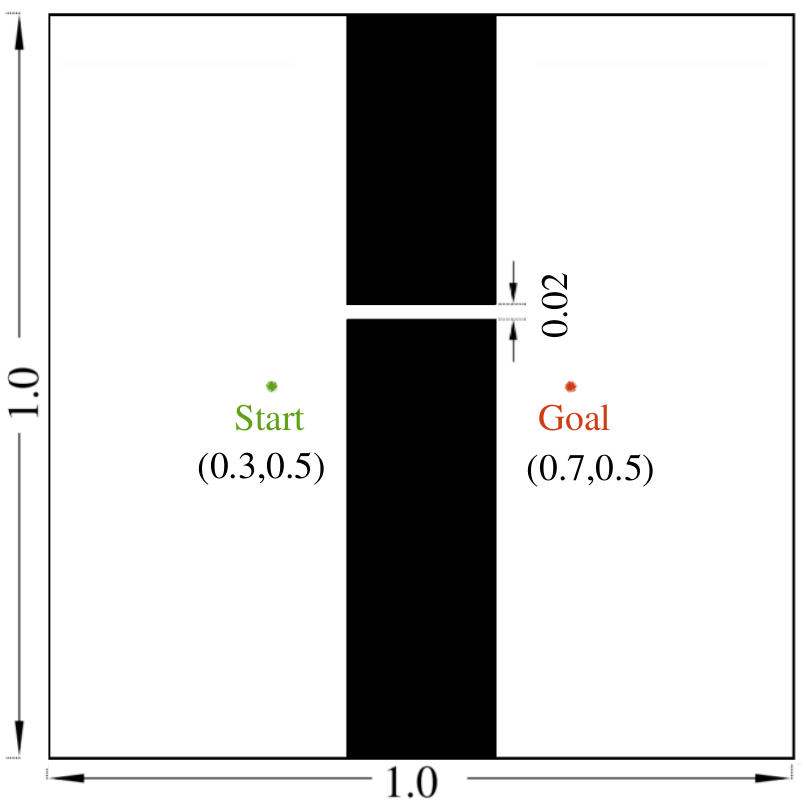}};
    \node[inner sep=0pt] (russell) at (0.25,0.0)
    {\includegraphics[width=0.24\textwidth]{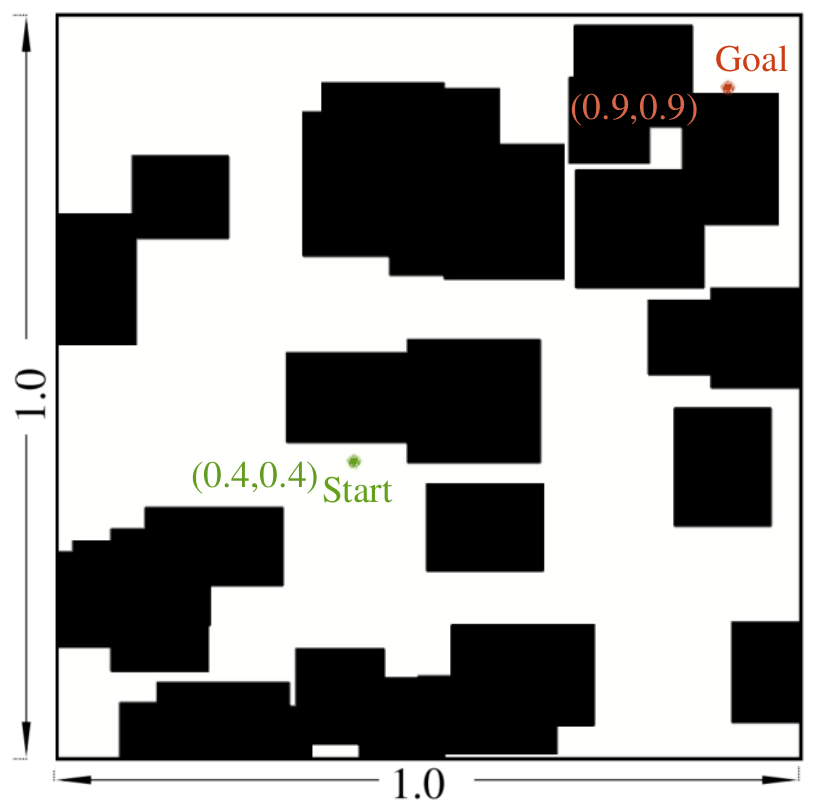}};
    \scriptsize
    \node at (-4.0,-2.51) {\small (a) Narrow Passage (NP)};
    \node at (0.25,-2.51) {\small (b) Random Rectangles (RR)};
    \end{tikzpicture}
    \caption{The 2D experimental scenarios.}
    \label{fig: testEnv}
    \vspace{-1.7em} 
\end{figure}

\subsection{Verification of Just-in-Time Module}\label{subsec:experi}

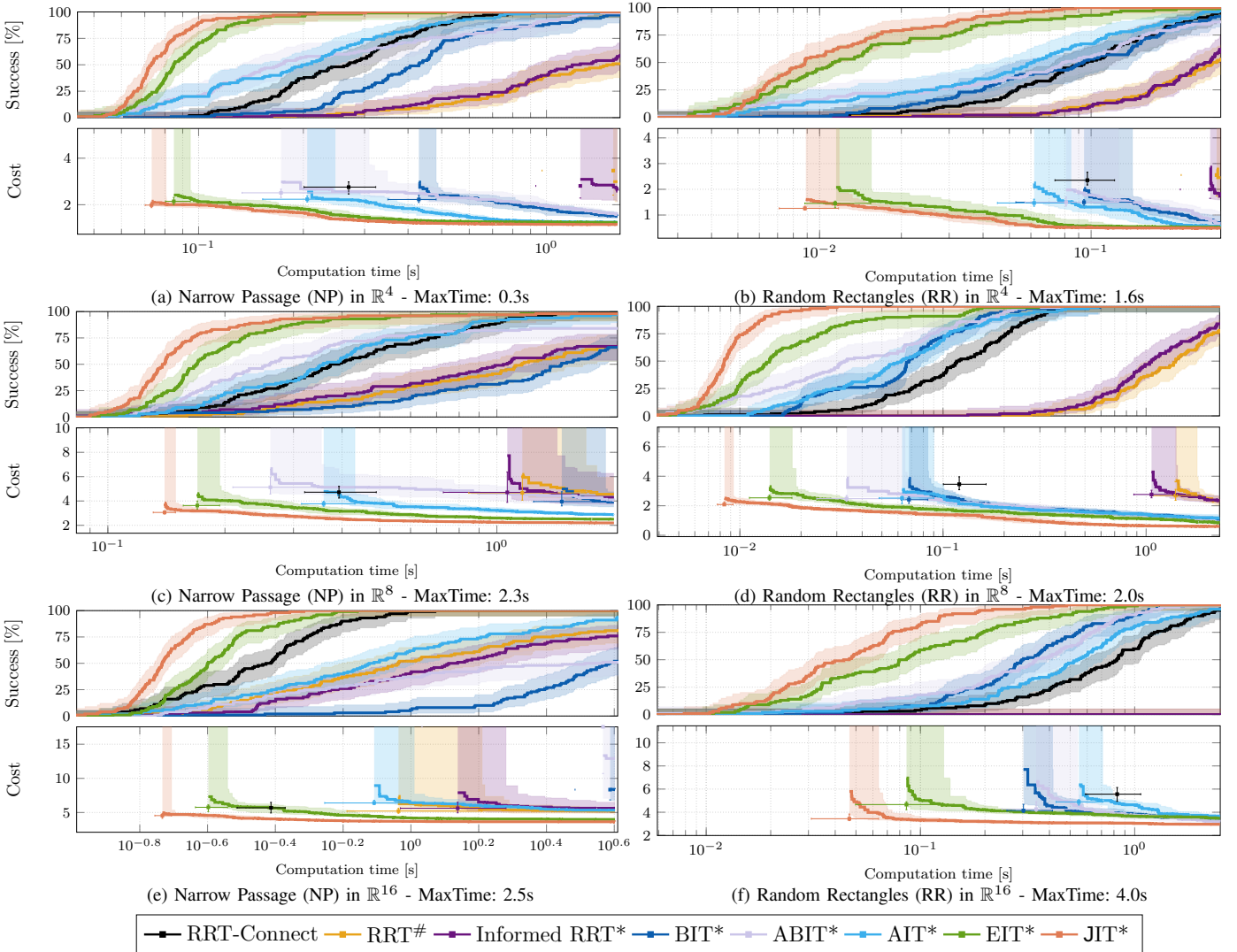
\begin{figure*}[t!]
    \centering
    \begin{tikzpicture}
    \node[inner sep=0pt] (russell) at (4.1,8)
    {\includegraphics[width=0.49\textwidth]{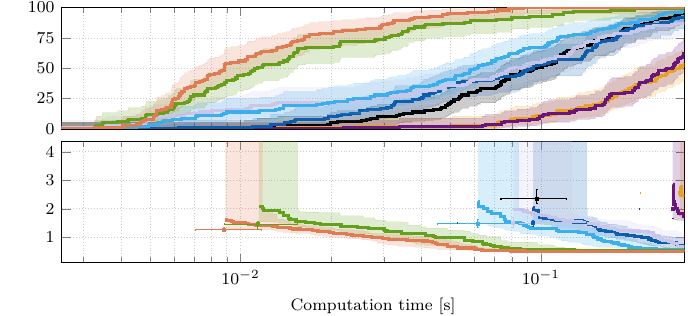}};
    \node[inner sep=0pt] (russell) at (4.1,3.5)
    {\includegraphics[width=0.49\textwidth]{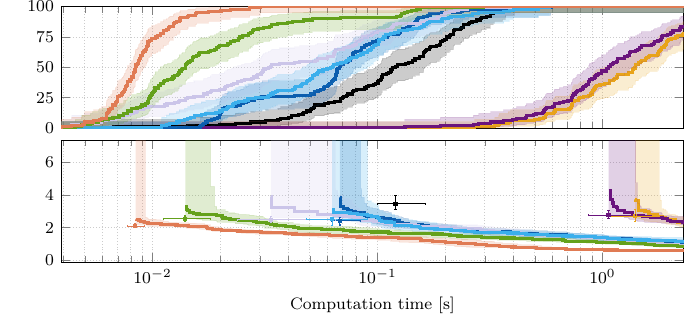}};
    \node[inner sep=0pt] (russell) at (4.1,-1)
    {\includegraphics[width=0.49\textwidth]{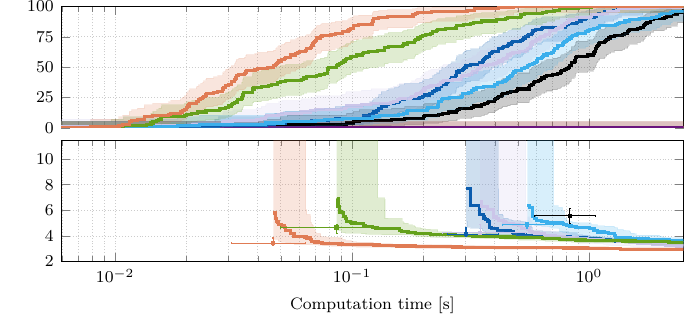}};

    \node[inner sep=0pt] (russell) at (-4.9,8)
    {\includegraphics[width=0.49\textwidth]{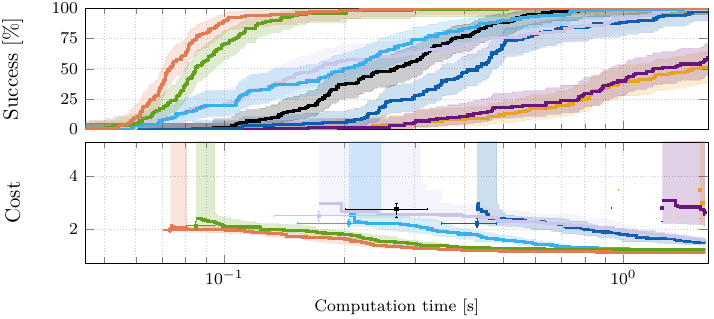}};  
    \node[inner sep=0pt] (russell) at (-4.93,3.5)
    {\includegraphics[width=0.489\textwidth]{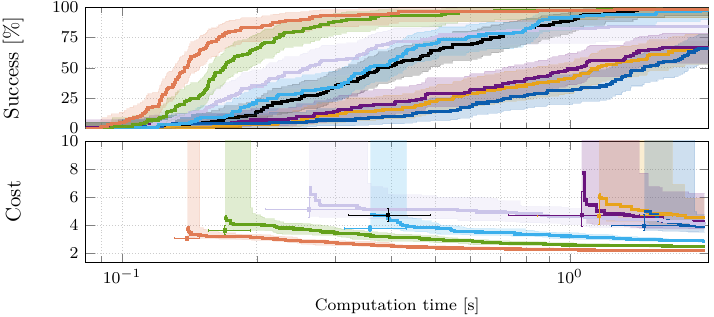}};
    \node[inner sep=0pt] (russell) at (-4.85,-1){\includegraphics[width=0.496\textwidth]{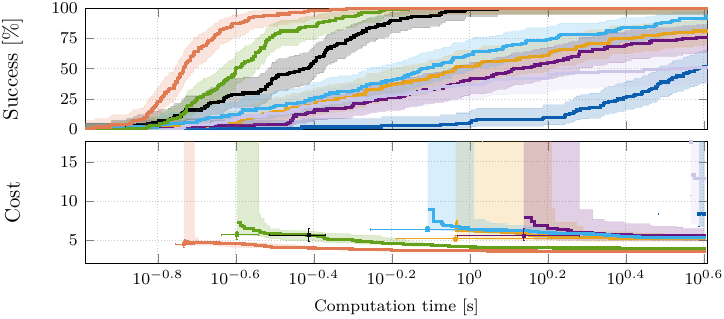}};

    \node[inner sep=0pt] (russell) at (0.0,-3.9){\includegraphics[width=0.8\textwidth]{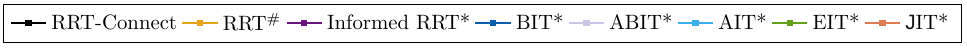}};

    \node at (-4.5,5.7) {\footnotesize (a) Narrow Passage (NP) in $\mathbb{R}^4$ - MaxTime: 0.3s};
    \node at (-4.5,1.2) {\footnotesize (c) Narrow Passage (NP) in $\mathbb{R}^8$ - MaxTime: 2.3s};
    \node at (-4.5,-3.3) {\footnotesize(e) Narrow Passage (NP) in $\mathbb{R}^{16}$ - MaxTime: 2.5s};

    \node at (4.5,5.7) {\footnotesize (b) Random Rectangles (RR) in $\mathbb{R}^4$ - MaxTime: 1.6s};
    \node at (4.5,1.2) {\footnotesize(d) Random Rectangles (RR) in $\mathbb{R}^8$ - MaxTime: 2.0s};
    \node at (4.5,-3.3) {\footnotesize(f) Random Rectangles (RR) in $\mathbb{R}^{16}$ - MaxTime: 4.0s};

    \end{tikzpicture}
    \vspace{-0.8em} 
    \caption{Experimental results for Just-in-time verification. MaxTime is the planner's maximum allotted planning time. (a), (c), and (e) depict test benchmark narrow passage outcomes in $\mathbb{R}4$, $\mathbb{R}^8$, and $\mathbb{R}^{16}$, respectively. (b), (d), and (f) showcase random rectangle experiments in $\mathbb{R}^4$, $\mathbb{R}^8$, $\mathbb{R}^{16}$.}
    \label{fig: benchmarkresult}
    \vspace{-1.8em}
\end{figure*}

To assess the feasibility of the Just-in-Time module, which optimizes edge expansion and sampling for efficient pathfinding, we compared our planner with and without this module against seven other methods in $\mathbb{R}^4$, $\mathbb{R}^8$, and $\mathbb{R}^{16}$ across two scenarios.
The first scenario (Fig.~\ref{fig: testEnv}a) features a narrow passage, where low sampling probability makes feasible path discovery difficult. The second scenario (Fig.~\ref{fig: testEnv}b) introduces randomly generated rectangular obstacles and feasible regions. Each planner was run 100 times, recording \textbf{success rate} (Success) and \textbf{path length} (Cost) within the maximum computation time (MaxTime), as shown in Fig.~\ref{fig: benchmarkresult} and Table~\ref{tab:benchmark}.

Table~\ref{tab:benchmark} shows that our algorithm with the Just-in-Time module consistently achieved the lowest values for $t^\textit{med}_\textit{init}$, $c^\textit{med}_\textit{init}$, and $c^\textit{med}_\textit{final}$, demonstrating its ability to rapidly find low-cost initial solutions. Notably, improvements in $T^\textit{med}_\textit{init}$, $C^\textit{med}_\textit{init}$, and $C^\textit{med}_\textit{final}$ over the second-best approach confirm its superior performance in success rate and cost throughout planning. This advantage stems from the Just-in-Time model, which reduces path costs by dynamically reconnecting vertices to form shorter edges without range constraints. Additionally, the Just-in-Time sampling model, particularly effective in NP-hard scenarios, improves success rates and $t^\textit{med}_\textit{init}$ by retaining collision-free edges and increasing sampling around failed edges to form new connections, enhancing overall planning success.

\textcolor{black}{Our just-in-time planner follows an almost-surely asymptotically optimal (ASAO) strategy, ensuring anytime performance by quickly finding an initial feasible solution and refining it toward optimality. As shown in Table~\ref{tab:benchmark}, it achieves short initial solution times with minimal cost across dimensions, maintaining the highest success rate and lowest cost throughout planning, making it ideal for real-time robotic applications.}

\begin{table*}[t]
\caption{Benchmarks evaluation comparison}
\centering
\resizebox{0.92\textwidth}{!}{
\begin{tabular}{p{1.5cm}||c|c|c||c|c|c||c|c|c||c|c|c|c}
 \hline
 & \multicolumn{3}{c||}{\textbf{RRT-Connect}} & \multicolumn{3}{c||}{\textbf{RRT\#}} & \multicolumn{3}{c||}{\textbf{Informed RRT*}} & \multicolumn{3}{c|}{\textbf{BIT*}} & \multirow{2}{*}{\textbf{$T^\textit{med}_\textit{init}$ (\%)}} \\
 & \( t^\textit{med}_\textit{init} \) & \( c^\textit{med}_\textit{init} \) & \( c^\textit{med}_\textit{final} \) & \( t^\textit{med}_\textit{init} \) & \( c^\textit{med}_\textit{init} \) & \( c^\textit{med}_\textit{final} \) & \( t^\textit{med}_\textit{init} \) & \( c^\textit{med}_\textit{init} \) & \( c^\textit{med}_\textit{final} \) & \( t^\textit{med}_\textit{init} \) & \( c^\textit{med}_\textit{init} \) & \( c^\textit{med}_\textit{final} \) & \\
 \hline
    \( \text{NP}-\mathbb{R}^4 \) & 0.096 & 2.353 & 2.353 & 0.290 & 2.555 & 2.452 & 0.274 & 1.996 & 1.731 & {0.094} & {1.497} & {0.691} & 18.18 \\
    \( \text{NP}-\mathbb{R}^8 \) & 0.635 & 5.031 & 5.031 & 3.221 & 3.274 & 2.896 & 2.710 & 3.175 & 2.762 & {0.377} & {2.689} & {1.945} & 75.76 \\
    \( \text{NP}-\mathbb{R}^{16} \) & 0.828 & 5.561 & 5.561 &$\infty$ &$\infty$ &$\infty$ & $\infty$ & $\infty$ & $\infty$ & {0.303} & \textcolor{teal}{4.147} & {3.484} & 45.35 \\
    \( \text{RR}-\mathbb{R}^4 \) & 0.270 & 2.758 & 2.758 & 1.556 & 3.468 & 2.991 & 1.251 & 2.809 & 2.582 & {0.173} & {2.227} & {1.508} & 14.12 \\
    \( \text{RR}-\mathbb{R}^8 \) & 0.120 & 3.457 & 3.457 & 1.397 & 2.690 & 2.344 & 1.064 & 2.764 & 2.344 & {0.068} & {2.412} & {1.003} & 17.65 \\
    \( \text{RR}-\mathbb{R}^{16} \) & 0.387 & 5.692 & 5.692 & {0.919} & \textcolor{teal}{5.206} & {5.206} & 1.374 & 5.621 & 5.621 & 3.871 & 8.347 & 8.347 & 26.88 \\
 \hline
  \hline
 & \multicolumn{3}{c||}{\textbf{ABIT*}} & \multicolumn{3}{c||}{\textbf{AIT*}} & \multicolumn{3}{c||}{\textbf{EIT*}} & \multicolumn{3}{c|}{\textbf{JIT*}} & \multirow{2}{*}{\textbf{$C^\textit{med}_\textit{init}$ / $C^\textit{med}_\textit{final}$ (\%)}} \\
 & \( t^\textit{med}_\textit{init} \) & \( c^\textit{med}_\textit{init} \) & \( c^\textit{med}_\textit{final} \) & \( t^\textit{med}_\textit{init} \) & \( c^\textit{med}_\textit{init} \) & \( c^\textit{med}_\textit{final} \) & \( t^\textit{med}_\textit{init} \) & \( c^\textit{med}_\textit{init} \) & \( c^\textit{med}_\textit{final} \) & \( t^\textit{med}_\textit{init} \) & \( c^\textit{med}_\textit{init} \) & \( c^\textit{med}_\textit{final} \) & \\
\hline
    \( \text{NP}-\mathbb{R}^4 \) & 0.081 & 1.498 & 0.671 & 0.062 & 1.469 & 0.559 & \textcolor{teal}{0.011} & \textcolor{teal}{1.453} & \textcolor{teal}{0.529} & \textcolor{red}{0.009} & \textcolor{red}{1.260} & \textcolor{red}{0.489} & 13.28 / 7.56 \\
    \( \text{NP}-\mathbb{R}^8 \) & 0.357 & \textcolor{teal}{2.646} & {1.982} & 0.500 & 2.717 & 2.043 & \textcolor{teal}{0.064} & 2.742 & \textcolor{teal}{1.898} & \textcolor{red}{0.027} & \textcolor{red}{2.178} & \textcolor{red}{1.488} & 17.69 / 21.60 \\
    \( \text{NP}-\mathbb{R}^{16} \) & {0.347} & {4.265} &  \textcolor{teal}{3.169} & 0.549 & 4.891 & 3.654 & \textcolor{teal}{0.086} & 4.663 & {3.486} & \textcolor{red}{0.047} & \textcolor{red}{3.439} & \textcolor{red}{2.963} & 17.07 / 5.91 \\
    \( \text{RR}-\mathbb{R}^4 \) & 0.173 & 2.512 & 1.583 & 0.205 & 2.236 & 1.248 & \textcolor{teal}{0.085} & \textcolor{teal}{2.146} & \textcolor{teal}{1.238} & \textcolor{red}{0.073} & \textcolor{red}{1.983} & \textcolor{red}{1.149} & 7.60 / 7.20 \\
    \( \text{RR}-\mathbb{R}^8 \) & \textcolor{teal}{0.033} & 2.479 & 0.901 & 0.063 & 2.518 & 1.152 &{0.014} & \textcolor{teal}{2.537} & \textcolor{teal}{0.847} & \textcolor{red}{0.008} & \textcolor{red}{2.086} & \textcolor{red}{0.601} & 13.52 / 29.04 \\
    \( \text{RR}-\mathbb{R}^{16} \) & 3.688 & 17.547 & 12.905 & 0.779 & 6.400 & 5.338 & \textcolor{teal}{0.253} & {5.762} & \textcolor{teal}{3.985} & \textcolor{red}{0.185} & \textcolor{red}{4.544} & \textcolor{red}{3.608} & 12.72 / 9.46 \\
 \hline
\end{tabular}}
\label{tab:benchmark}
\vspace{-1.3em}
\end{table*}

\begin{figure*}[h]
    \centering
    \subfloat[]{\includegraphics[width=0.16\textwidth]{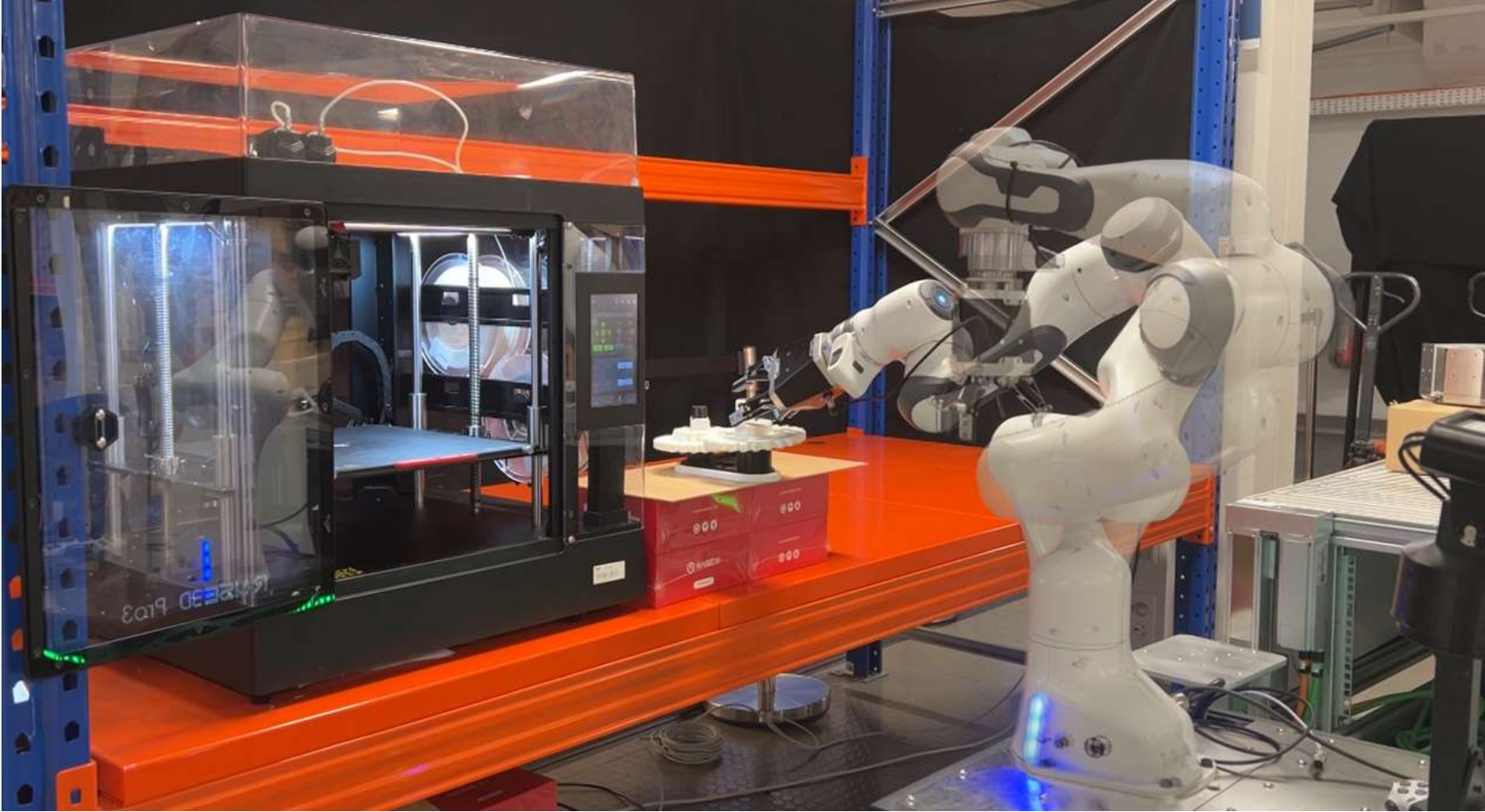}}
    \hfill
    \subfloat[]{\includegraphics[width=0.153\textwidth]{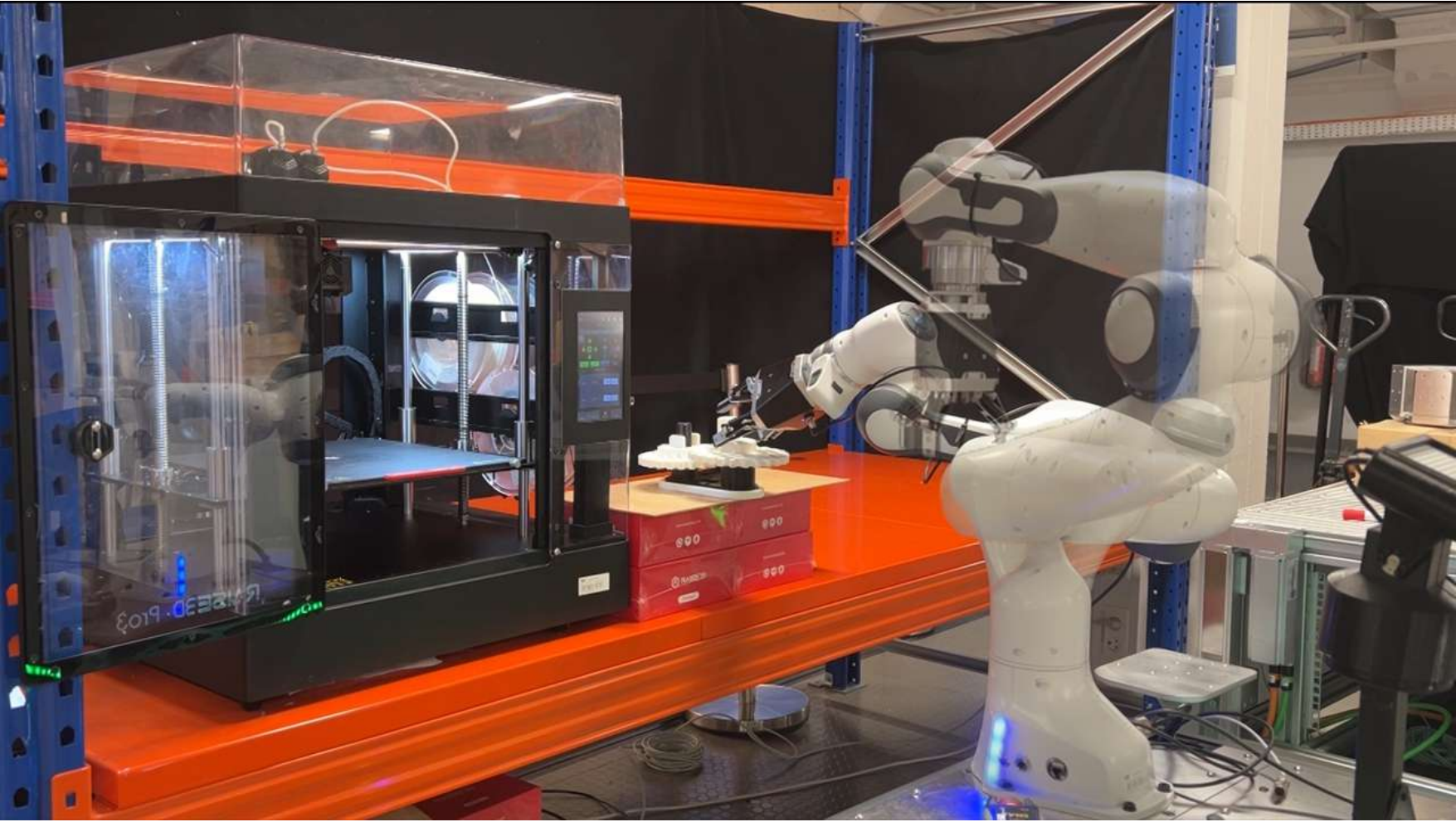}}
    \hfill
    \subfloat[]{\includegraphics[width=0.168\textwidth]{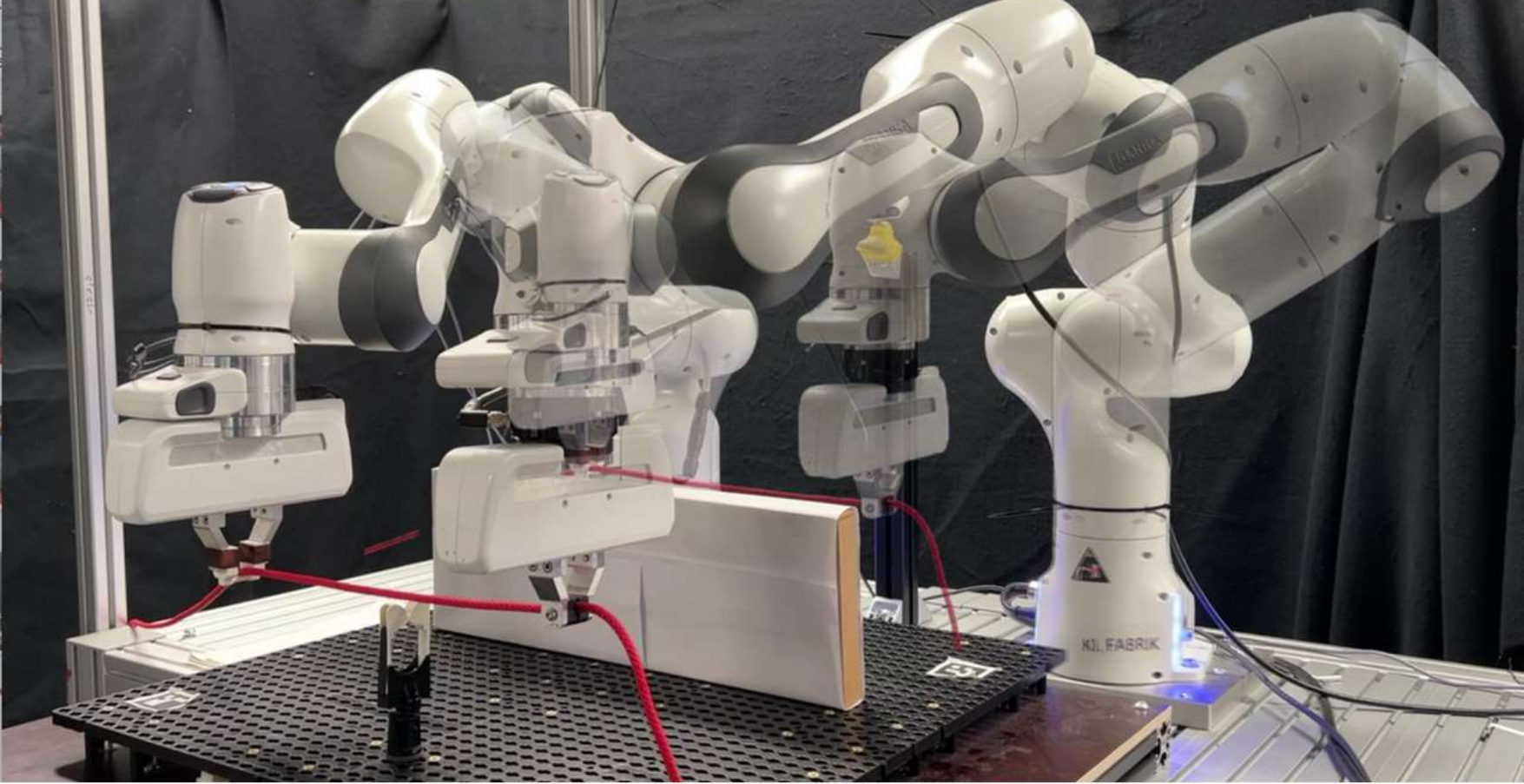}}
    \hfill
    \subfloat[]{\includegraphics[width=0.168\textwidth]{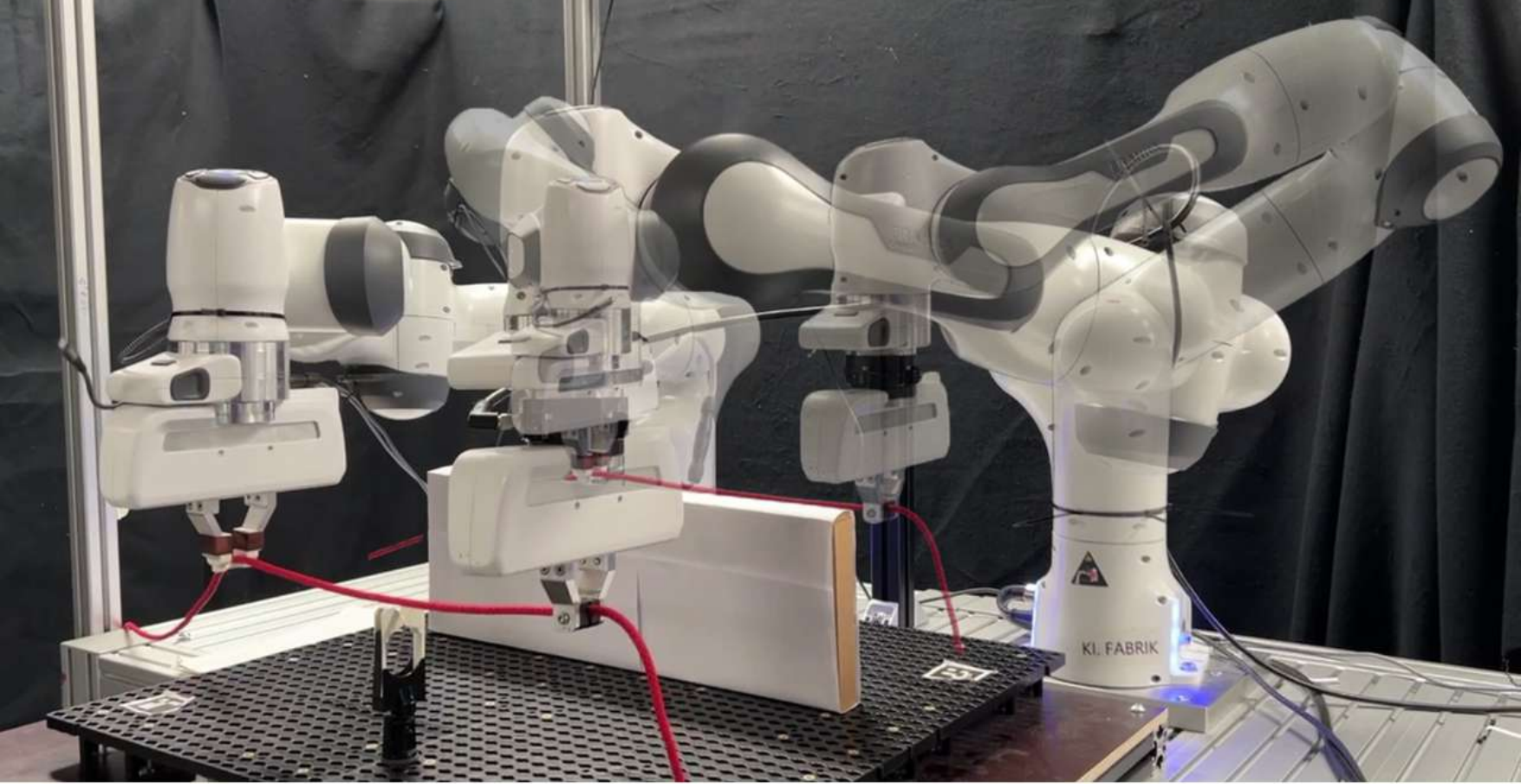}}
    \hfill
    \subfloat[]{\includegraphics[width=0.16\textwidth]{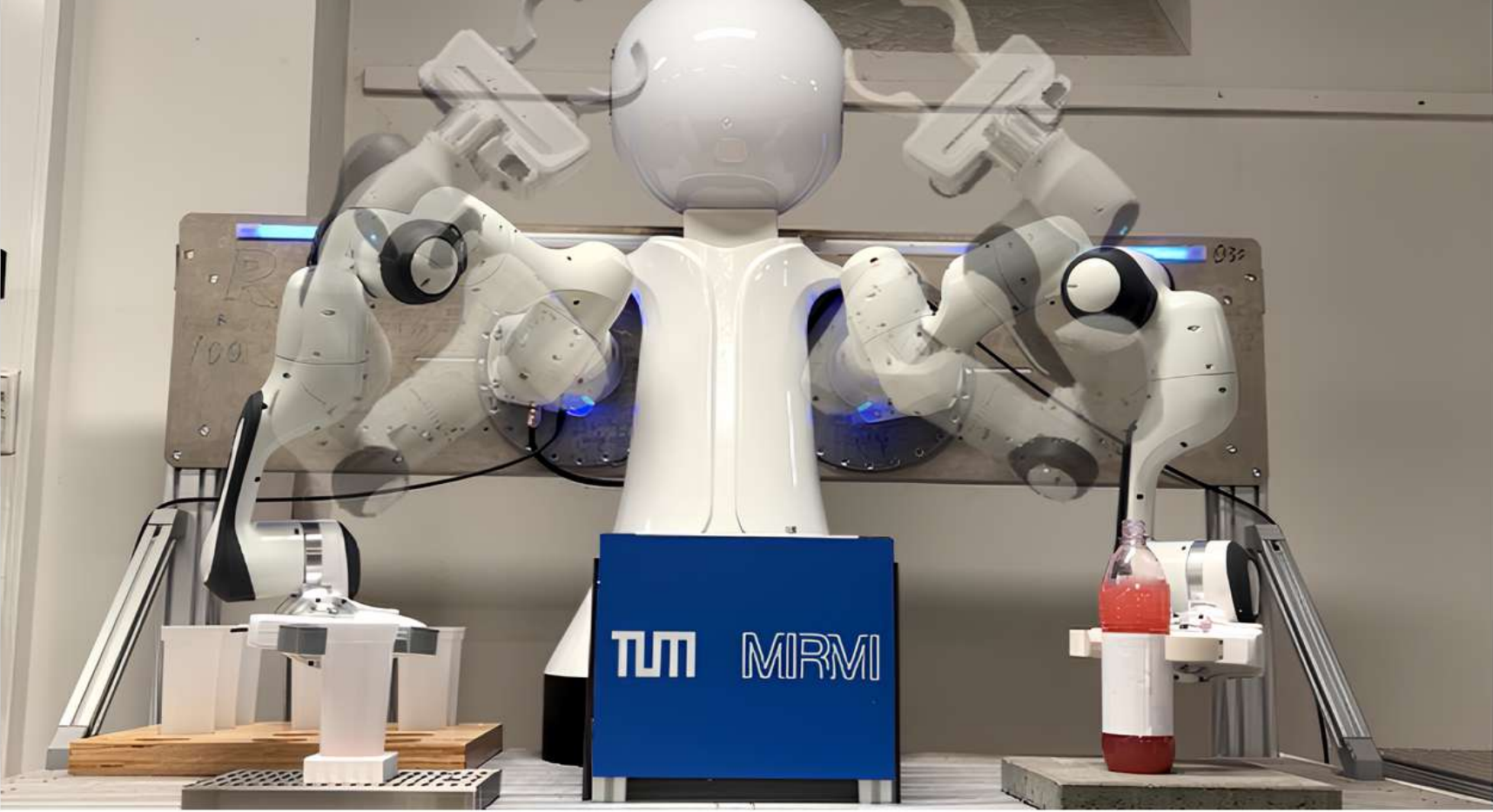}}
    \hfill
    \subfloat[]{\includegraphics[width=0.16\textwidth]{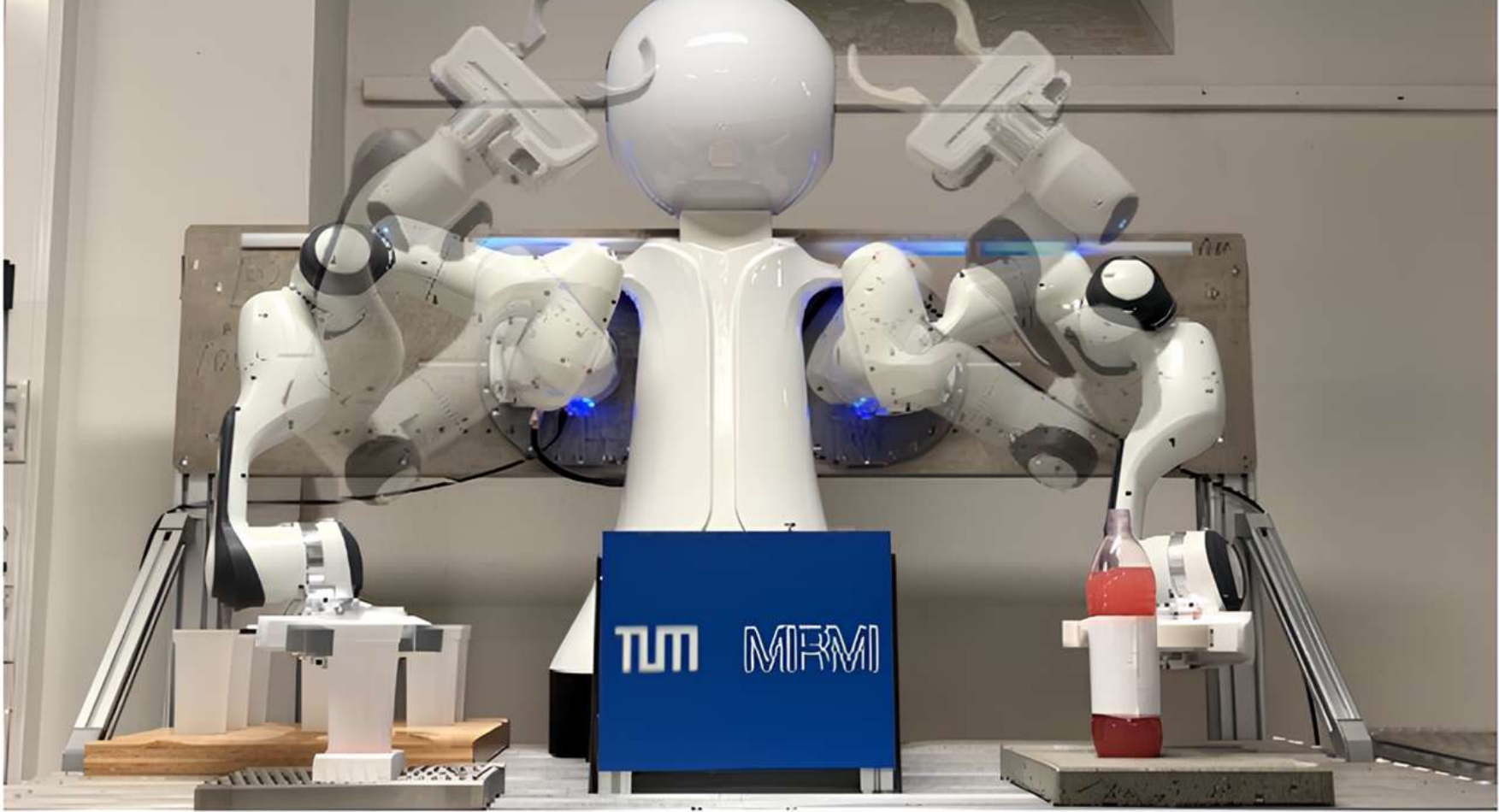}}
    \vspace{0.01em} 
    \subfloat[]{\includegraphics[width=0.19\textwidth]{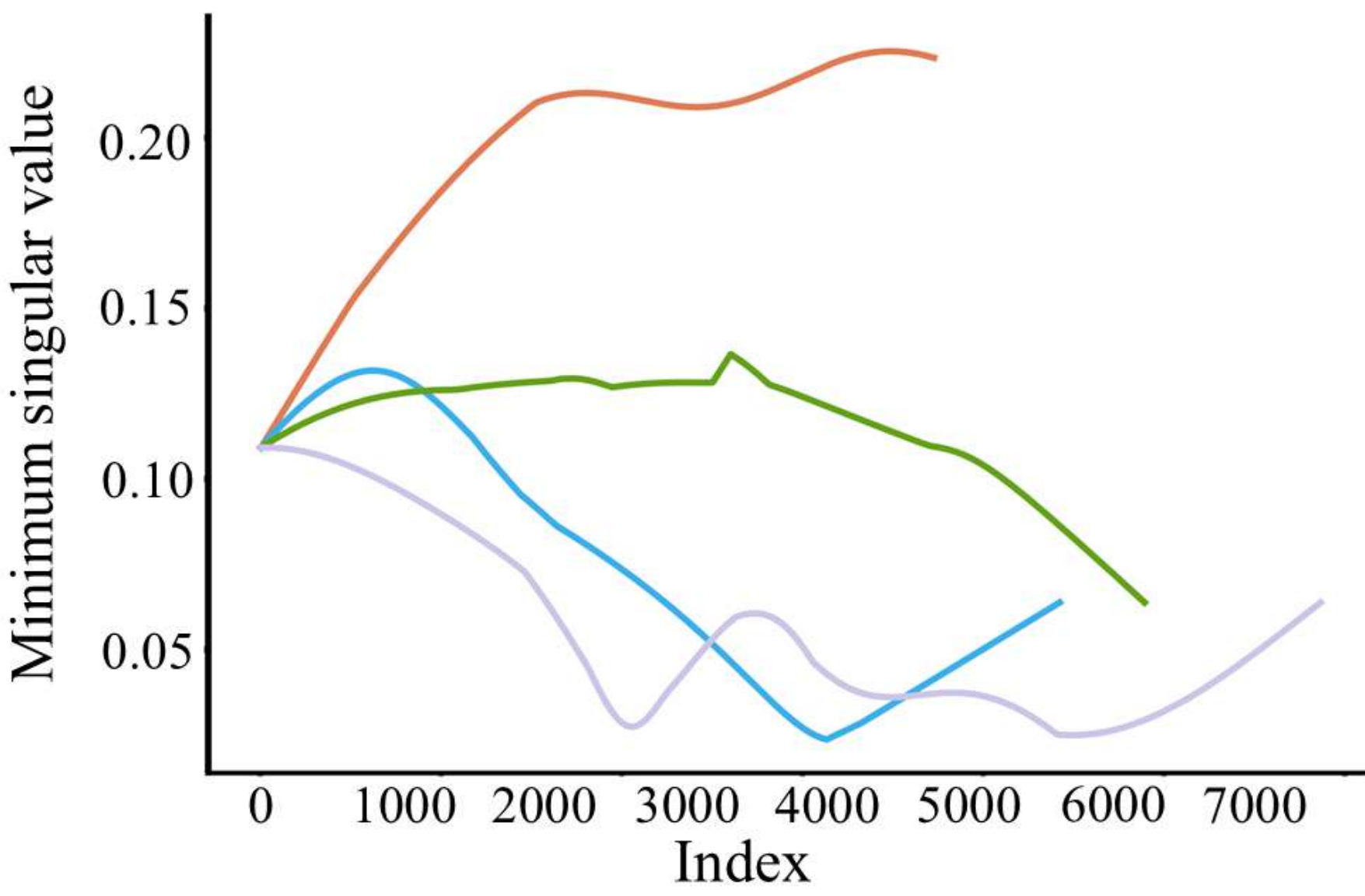}}
    \hfill
    \subfloat[]{\includegraphics[width=0.19\textwidth]{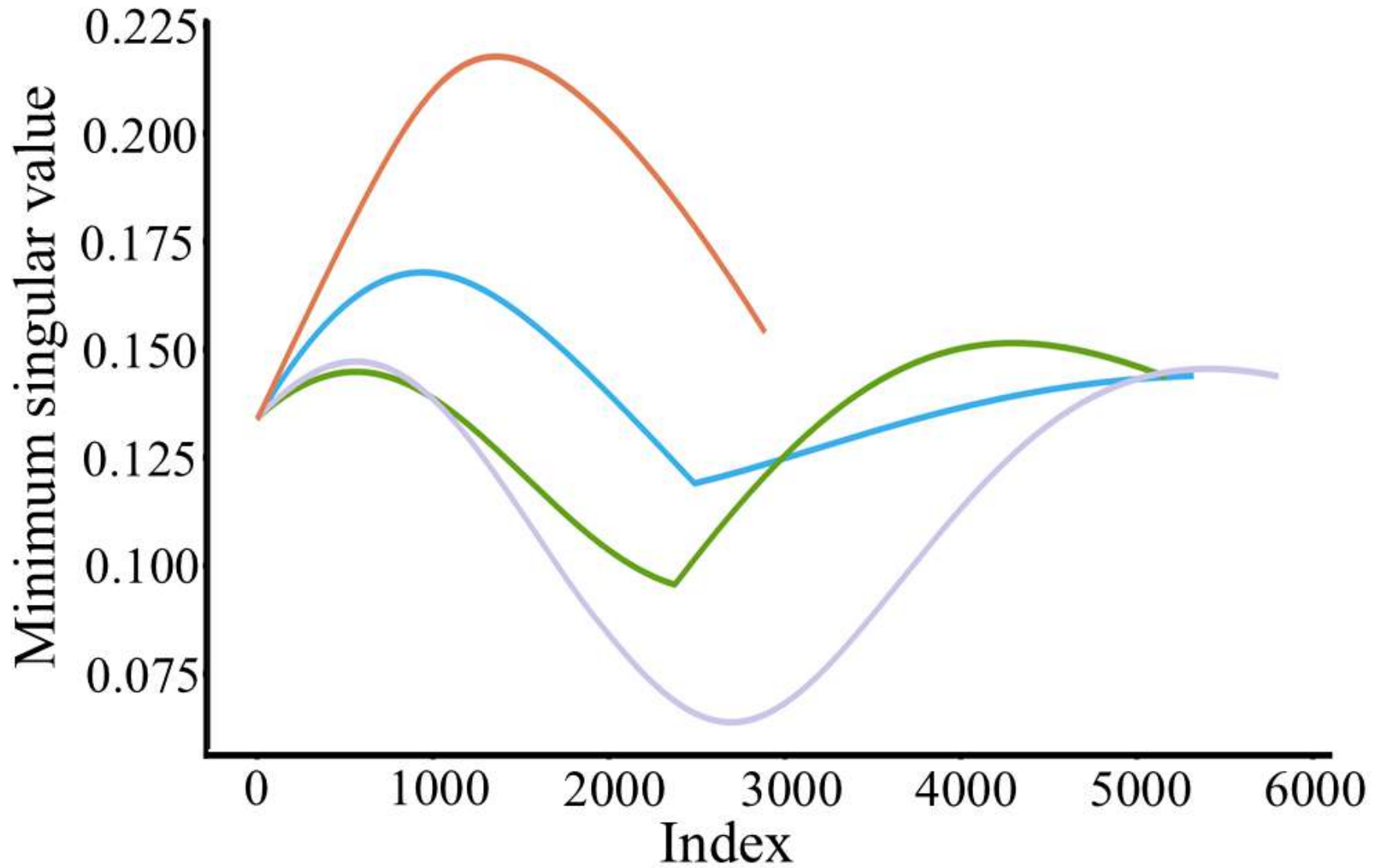}}
    \hfill
    \subfloat[]{\includegraphics[width=0.2\textwidth]{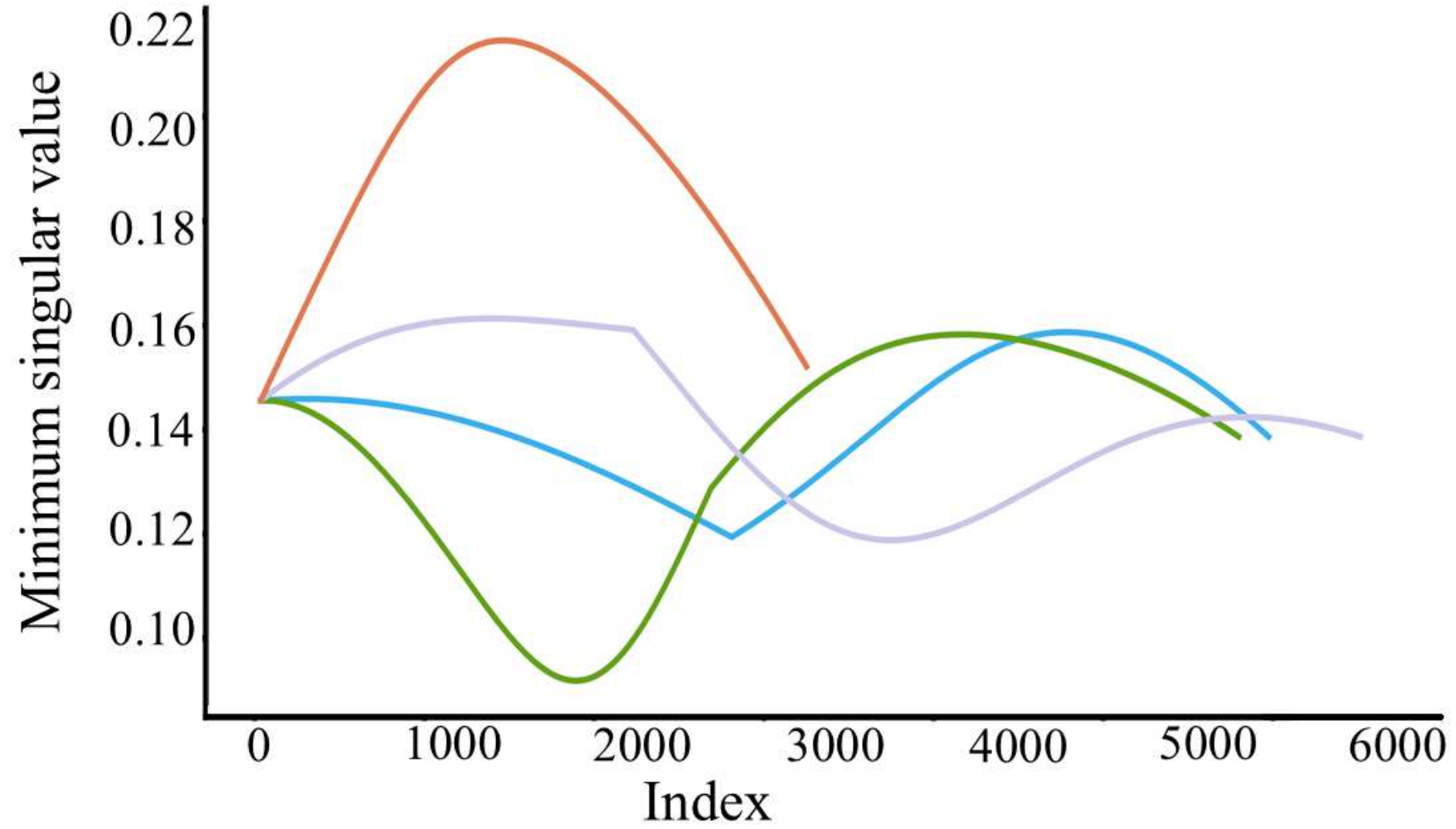}}
    \hfill
    \subfloat[]{\includegraphics[width=0.195\textwidth]{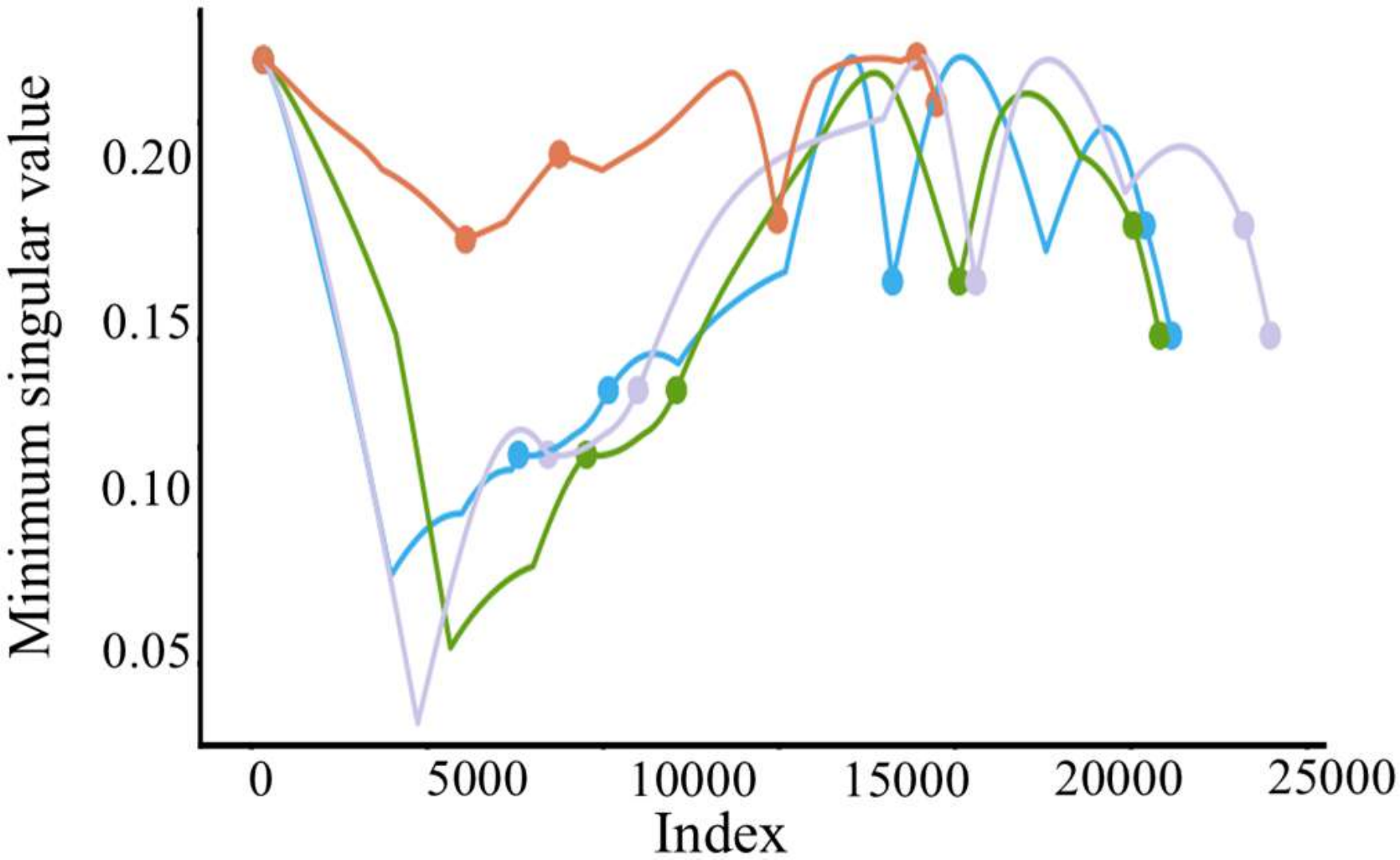}}
    \hfill
    \subfloat[]{\includegraphics[width=0.2\textwidth]{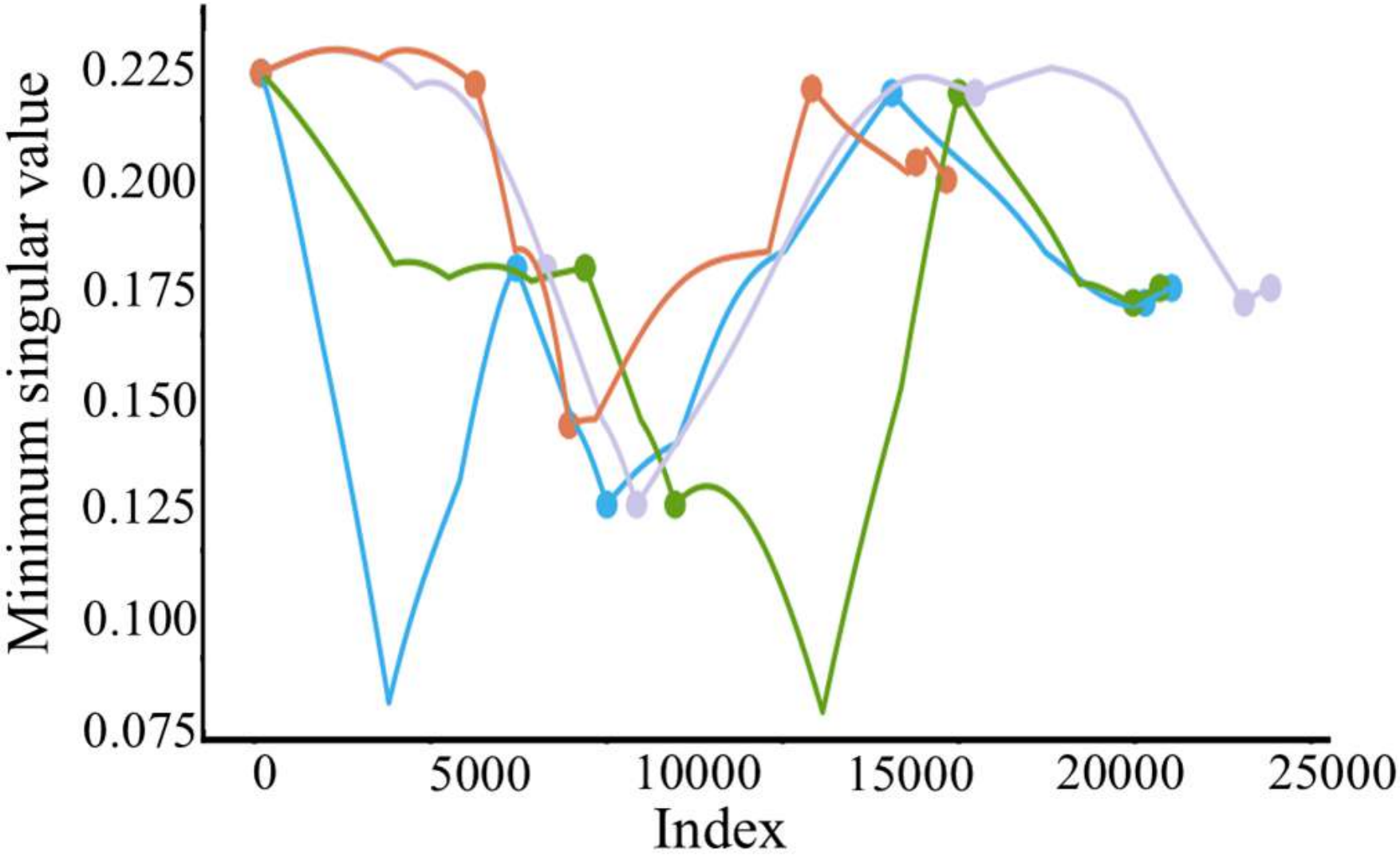}}
    \vspace{0.01em} 
    \subfloat[]{\includegraphics[height = 0.16\textwidth,width=0.16\textwidth]{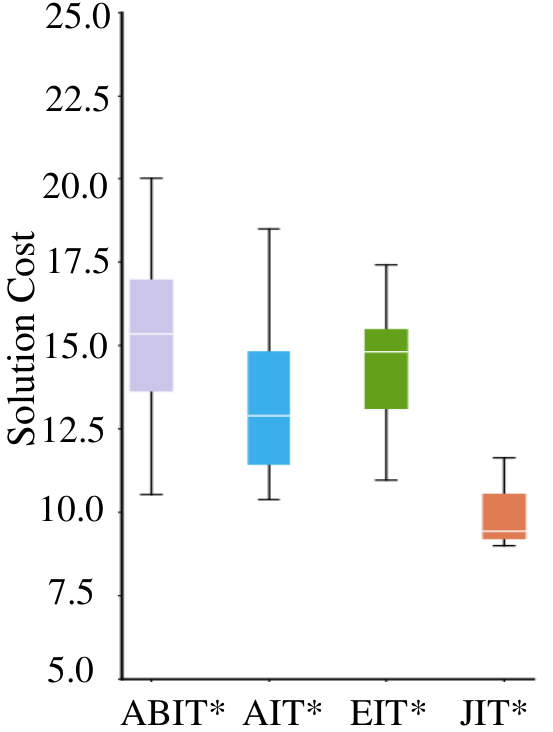}}
    \hfill
    \subfloat[]{\includegraphics[height = 0.16\textwidth,width=0.16\textwidth]{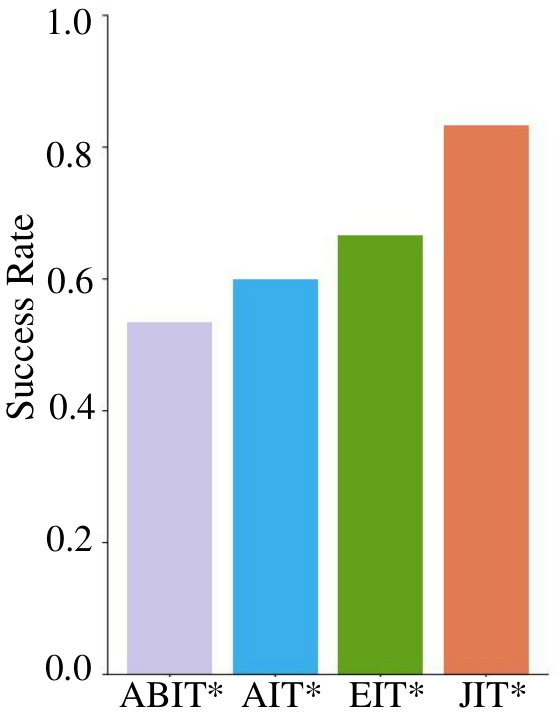}}
    \hfill
    \subfloat[]{\includegraphics[height = 0.16\textwidth,width=0.16\textwidth]{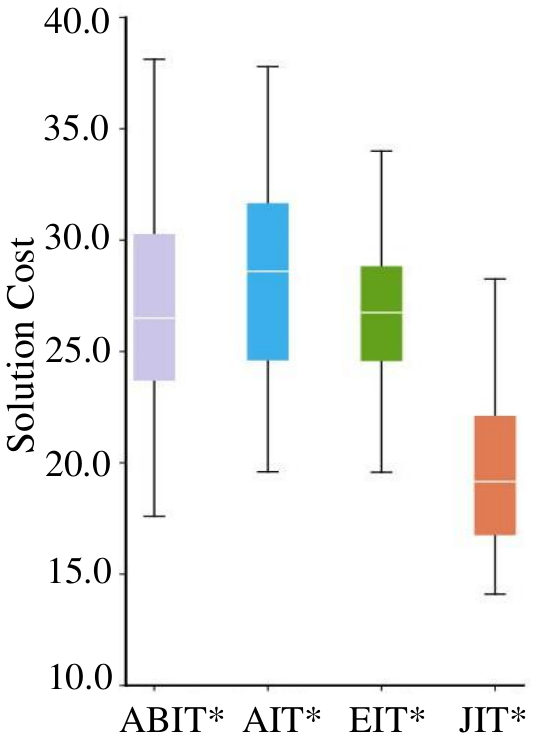}}
    \hfill
    \subfloat[]{\includegraphics[height = 0.16\textwidth,width=0.16\textwidth]{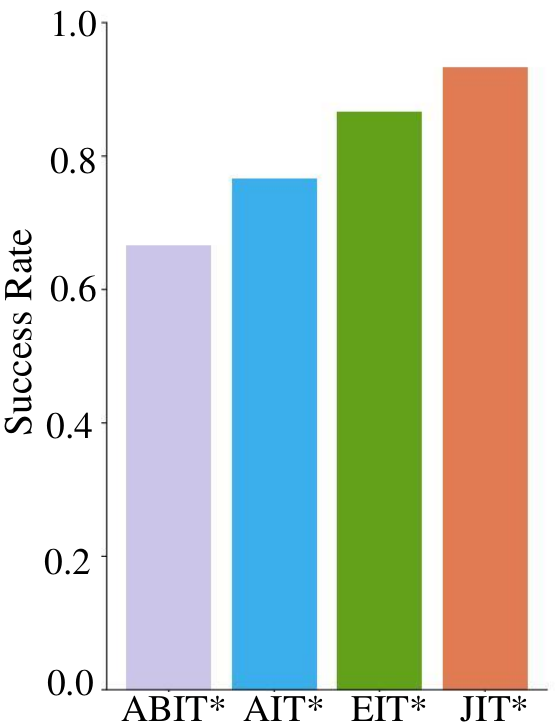}}
    \hfill
    \subfloat[]{\includegraphics[height = 0.16\textwidth,width=0.16\textwidth]{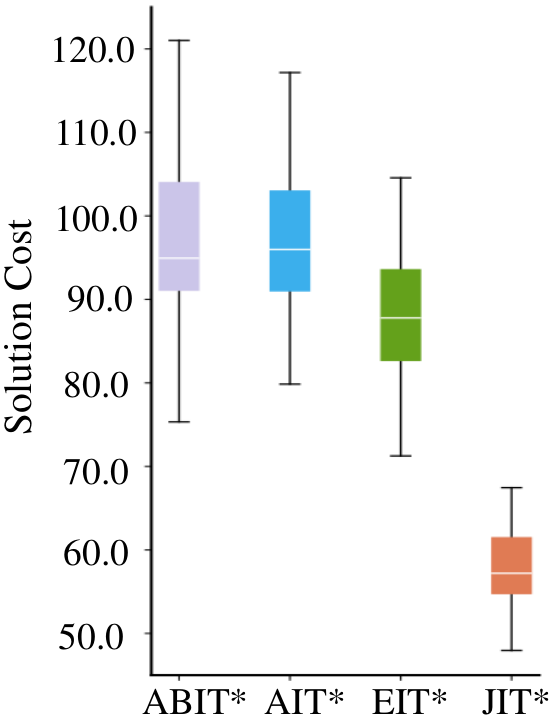}}
    \hfill
    \subfloat[]{\includegraphics[height = 0.16\textwidth,width=0.16\textwidth]{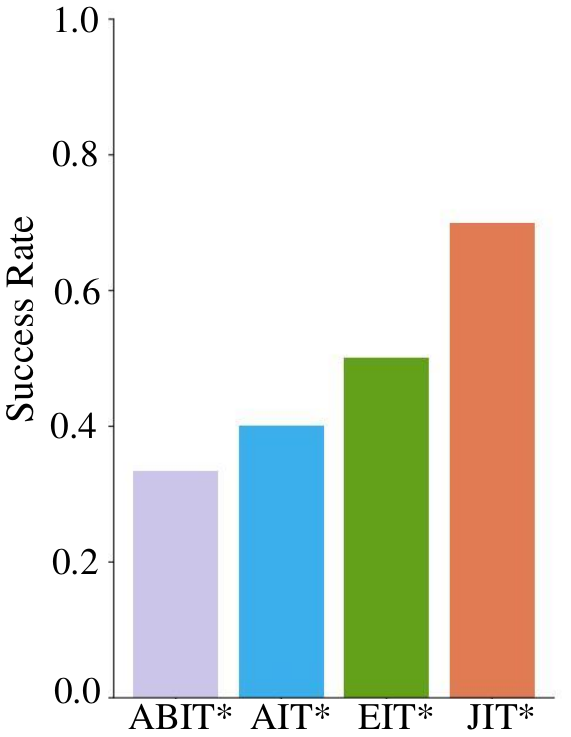}}
    \caption{Results of real robot experiments. Fig. (a) shows the GPIT start and end points, while Fig. (b) shows the optimized end configuration by our method with the same end-effector position but a different joint setup; dashed outlines indicate start positions, and solid outlines the end. Figs. (c) and (d) illustrate the DOPOT start and end points, and Figs. (e) and (f) display the PWT. Fig. (g) gives the minimal singularity value for the GPIT, with Figs. (h) and (j) showing minimal values for the right arm in DOPOT and PWT, and Figs. (i) and (k) for the left arm. The third row shows additional results: Figs. (l) and (m) show solution cost and success rate for GPIT, Figs. (n) and (o) for DOPOT, and Figs. (p) and (q) for PWT.}
    \label{fig:real-mani}
\end{figure*}

\subsection{Verification of Just-in-time module and motion performance module}\label{subsec:realExpri}

To evaluate the Just-in-Time and motion performance modules, we conducted three real-world experiments: \textit{Gear Pick and Insertion}, \textit{Deformable Object Pre-Clip Operation}, and \textit{Pour Water}. These included a single-arm scenario ($\mathbb{R}^7$) and two dual-arm scenarios ($\mathbb{R}^{14}$). Our method was compared with ABIT*, AIT*, and EIT*, each tested 30 times per scenario. Metrics included \textbf{optimal solution cost} and \textbf{success rate} for efficiency and \textbf{minimum singular value} for motion performance.


\subsubsection{\textbf{Gear pick and insertion task (GPIT)}}

To validate obstacle avoidance and precise end-effector positioning, we designed the Gear Pick and Insertion task. Fig.~\ref{fig:real-mani}(a) and (b) show the initial and final configurations. The robotic arm retrieves a printed gear and inserts it into an assembly within a confined space, avoiding collisions with the 3D printer. Precise alignment ensures the gear’s center matches the shaft and meshes correctly, requiring exact end-effector positioning.

Fig.\ref{fig:real-mani}(g) and Fig.\ref{fig:real-mani}(l) present performance results in manipulability and solution cost. Across 30 trials, our planner achieved the lowest median solution cost (58.09), outperforming EIT* (88.64), AIT* (96.20), and ABIT* (96.77). Tracking the minimum singularity value at each configuration, Fig.\ref{fig:real-mani}(g) shows our motion performance module consistently avoids singularities. As shown in Fig.\ref{fig:real-mani}(m), our method also achieves the highest success rate by accounting for self-collision, improving planning robustness.



\subsubsection{\textbf{Deformable object pre-clip operation task (DOPOT)}}

To evaluate the algorithm's performance in high-dimensional spaces, we designed a dual-arm deformable object pre-clip operation task. In this task, both robotic arms collaborate to manipulate a deformable rope, avoiding obstacles and placing it onto a clip. The initial and final configurations for the task are depicted in Fig.~\ref{fig:real-mani}(c) and (d). 
Our algorithm successfully enables the arms to pick up the rope and maneuver it around the white obstacle with minimal cost. The total cost for both arms combined is 19.0, significantly outperforming EIT* (26.68), AIT* (28.19), and ABIT* (27.17). Additionally, as shown in Fig. ~\ref{fig:real-mani}(h) and (i), the manipulability of both the left and right arms is higher than other algorithms. Moreover, our algorithm achieved a success rate of 93.33\%, surpassing EIT* (86.67\%), AIT* (76.67\%), and ABIT* (66.67\%) in Fig.~\ref{fig:real-mani}(o), demonstrating that it allows the robot to safely complete dual-arm obstacle-avoidance tasks.

\subsubsection{\textbf{Pour water task (PWT)}}

To further validate our algorithm’s efficiency and motion performance, we designed the Pour Water Task, comprising five subtasks: (1) approaching pre-grasp, (2) grasping and lifting, (3) moving to pre-pour, (4) pouring and positioning near the coaster, and (5) lowering and releasing. As shown in Fig. 4, our algorithm achieves 52.54\% lower cost than the second-best EIT*, driven by consistently superior performance across all subtasks. Figs.~\ref{fig:real-mani} (j) and (k) illustrate our algorithm’s ability to optimize manipulability at each subtask. It maintains minimum singularity values of 0.145 (left arm) and 0.176 (right arm), outperforming EIT* (0.079, 0.053), AIT* (0.082, 0.075), and ABIT* (0.127, 0.031). This improvement ensures smoother, more dexterous movements, reducing singularities and enhancing control efficiency.
For complex dual-arm tasks, our method significantly outperforms others in success rate by accounting for both single-arm and inter-arm collisions, as shown in Fig.~\ref{fig:real-mani}(q). JIT* achieves 70\% success, surpassing EIT*’s 56.7\% by 23.5\%. Overall, JIT* delivers the best efficiency and motion performance compared to EIT*, AIT*, and ABIT*.

\textcolor{black}{The JIT* algorithm excels in manipulability optimization and dynamic adaptability. By leveraging the null space, it continuously refines joint configurations while maintaining end-effector stability and avoiding singularities. Its scalability enables it to efficiently tackle a diverse range of tasks while ensuring high success rates. The Just-in-Time module enhances edge connectivity and sampling, facilitating rapid solution discovery and continuous adaptation in complex environments. Experimental results demonstrate that JIT* achieves superior success rates and manipulability compared to state-of-the-art methods, highlighting its efficiency in motion planning across different dimensions.}
\section{Conclusion}
In this paper, we presented the Just-in-Time Informed Trees (JIT*) algorithm, an advancement in path planning designed to efficiently navigate high-dimensional environments while optimizing robotic arm motion. JIT* addresses existing challenges by integrating two main modules: the Just-in-Time module, which dynamically enhances edge connectivity and sample density for rapid initial pathfinding, and the Motion Performance module, which balances manipulability and trajectory cost to avoid kinematic singularities and self-collisions. 
Comparative analyses show that JIT* consistently outperforms traditional sampling-based planners across problems with different dimensions, maintaining high efficiency and solution quality. In practical application, JIT* has demonstrated robust performance in both single-arm and dual-arm manipulation tasks, validating its effectiveness for real-world robotic planning.

\textcolor{black}{While JIT* enhances path planning efficiency and manipulability, its performance may decline in highly cluttered or dynamic environments, and its reliance on manipulability heuristics may not generalize well to tasks with other critical constraints like force control or compliance. In addition, although JIT* proves effective in up to 16-dimensional spaces, its computational complexity becomes restrictive as dimensionality grows, primarily due to exponential increases in sampling complexity and collision-checking overhead. Future work may explore more efficient sampling strategies \cite{zhang2024flexible} and hardware acceleration \cite{Wilson2024fcit}.}



\bibliographystyle{IEEEtran}
\bibliography{bibliography}
\begin{IEEEbiography}[{\includegraphics[width=1in,height=1.25in, clip,keepaspectratio]{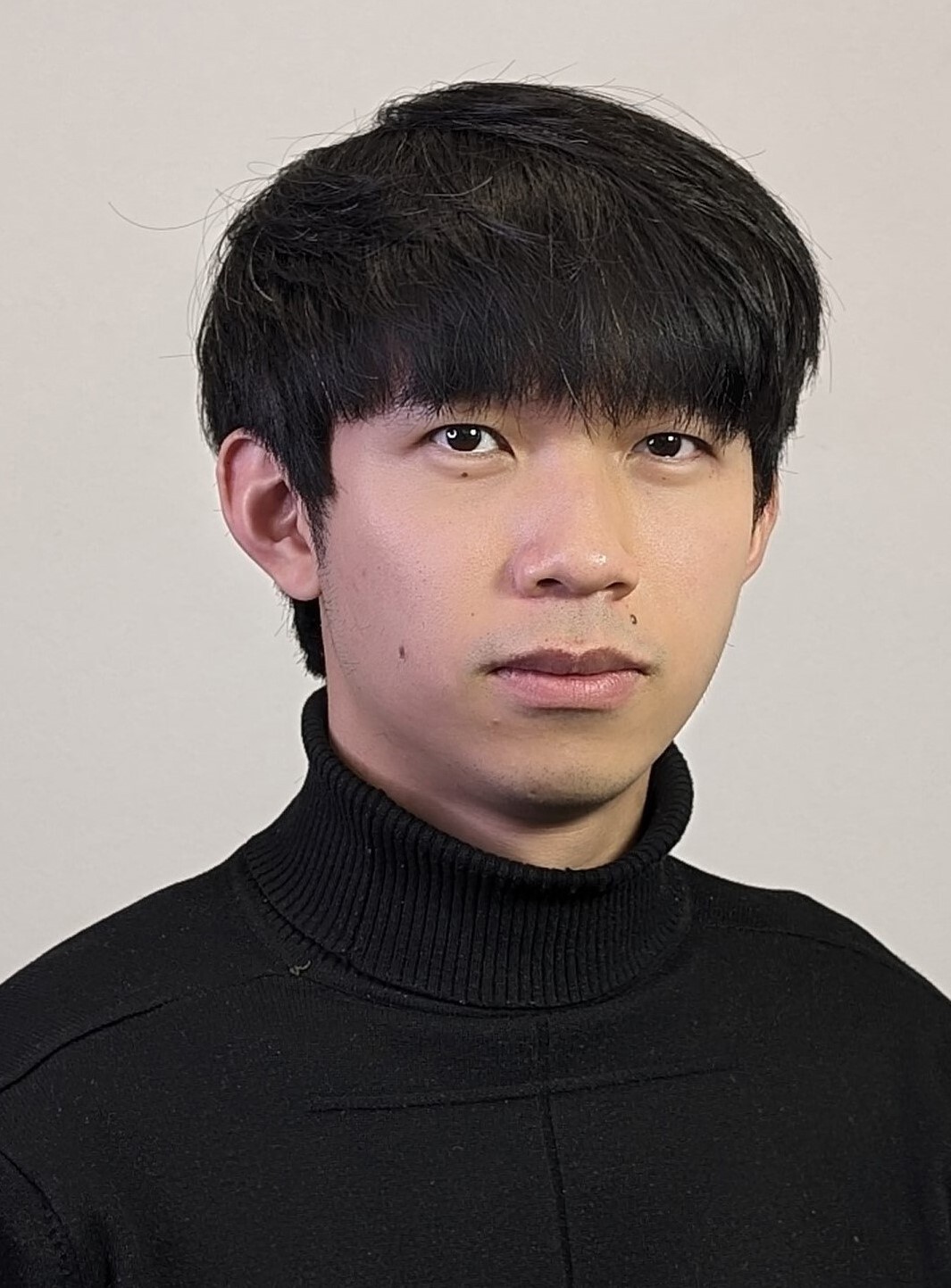}}] {Kuanqi Cai} is currently a Ph.D. student fellow at the HRII Lab, Italian Institute of Technology (IIT), and the LASA Lab, Swiss Federal Technology Institute of Lausanne (EPFL). Prior to joining IIT and EPFL, he was a Research Associate at the Technical University of Munich, from 2023 to 2024. 
In 2021, he was honored as a Robotics Student Fellow at ETH Zurich. 
He obtained his B.E. degree from Hainan University in 2018 and his M.E. degree from Harbin Institute of Technology in 2021. His current research interests focus on motion planning and human-robot interaction.

\end{IEEEbiography}

\vspace{11pt}

\begin{IEEEbiography}[{\includegraphics[width=1in,height=1.25in,clip,keepaspectratio]{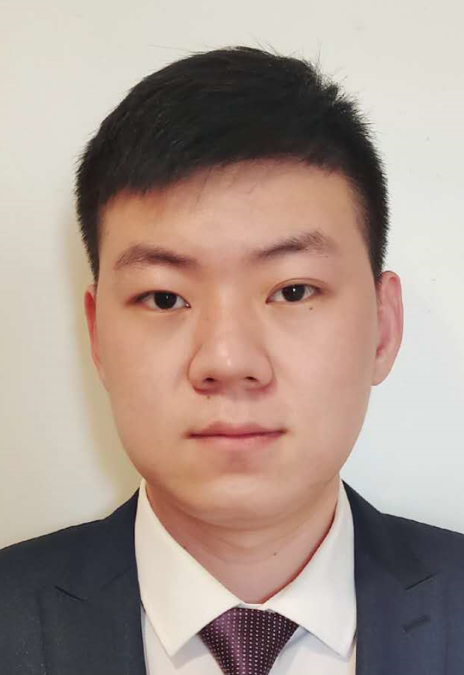}}] {Liding Zhang}is currently a Ph.D. candidate at the
School of Computation, Information and Technology (CIT) chair of informatics 6, Technical University of Munich, Germany. He received a B.Sc. degree in mechanical engineering from the Rhine-Waal University of Applied Sciences, Germany, in 2020 and an M.Sc. degree in mechanical engineering and automation from the Technical University of Clausthal, Germany, in 2022; his current research interests include robotic task and motion planning.
\end{IEEEbiography}

\vspace{11pt}

\begin{IEEEbiography}[{\includegraphics[width=1in,height=1.25in,clip,keepaspectratio]{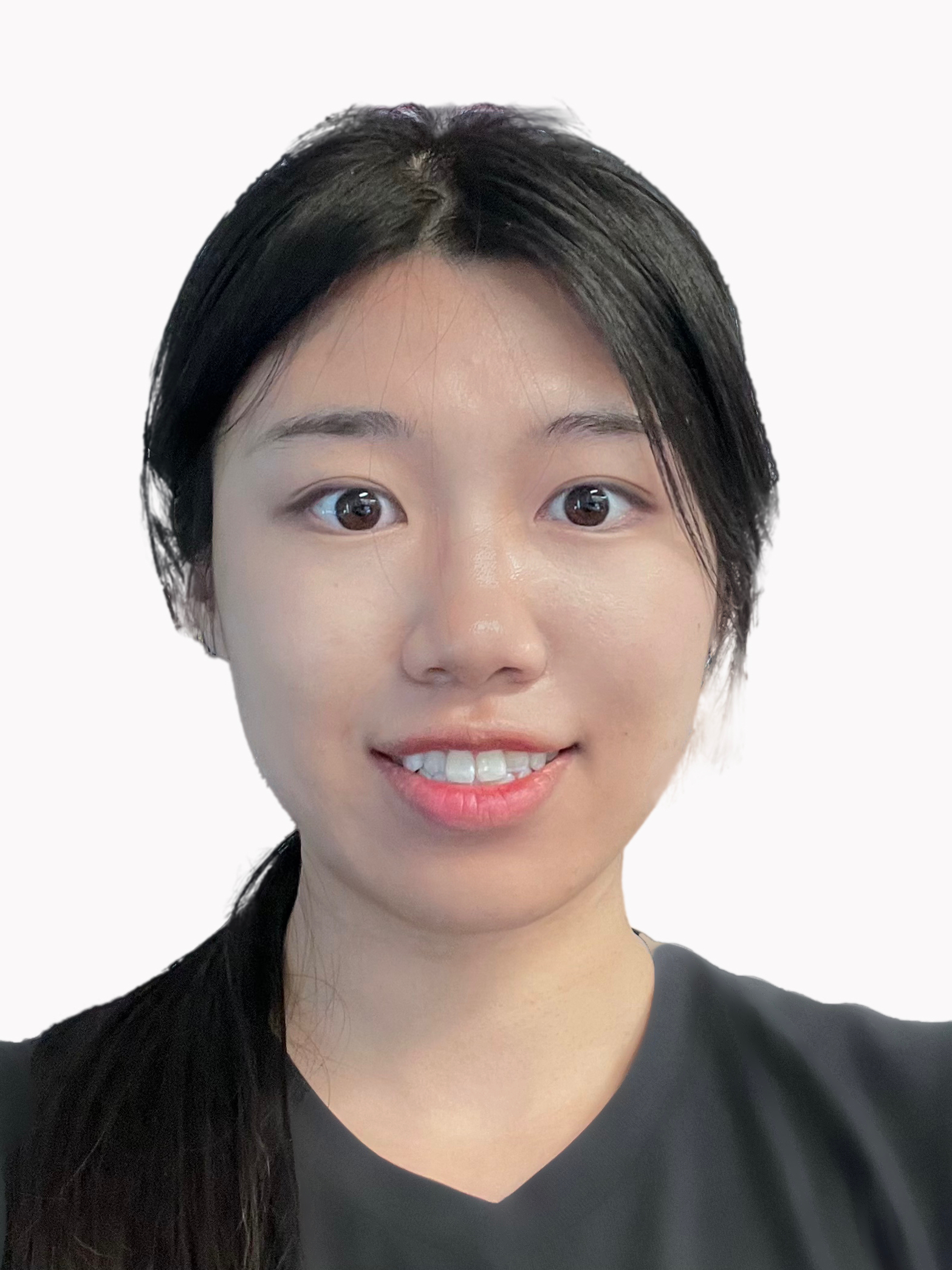}}] {Xinwen Su} is currently an Autonomous Driving Algorithm Engineer at Momenta. She received her M.Sc. degree in Mechatronics and Robotics from the Technical University of Munich (TUM), Germany, in 2024, and her B.Eng. degree in Vehicle Engineering from the Harbin Institute of Technology (HIT), China, in 2019. Her research interests include decision-making and planning for both autonomous driving systems and robotic arm systems.
\end{IEEEbiography}

\vspace{11pt}

\begin{IEEEbiography}[{\includegraphics[width=1in,height=1.25in,clip,keepaspectratio]{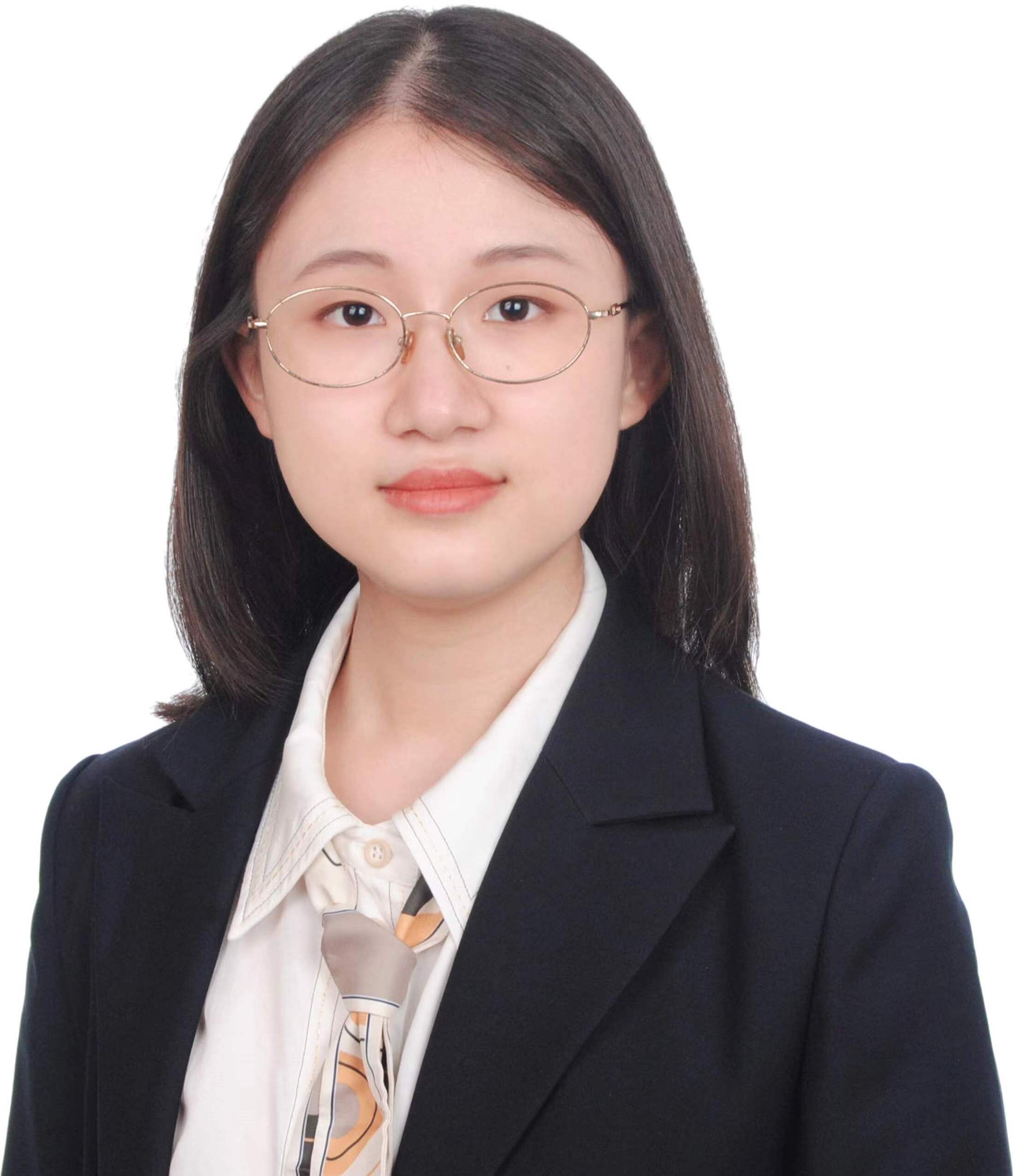}}] {Kejia Chen} is a doctoral candidate at the Chair of Robotics, Artificial Intelligence, and Real-Time Systems at the Technical University of Munich (TUM), Germany. She completed her M.Sc. in Robotics, Cognition, and Intelligence at TUM in 2021. Her current research interests include manipulation of deformable objects and multi-robot collaboration.
\end{IEEEbiography}

\vspace{11pt}

\begin{IEEEbiography}[{\includegraphics[width=1in,height=1.25in,clip,keepaspectratio]{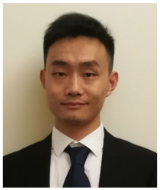}}]{Chaoqun Wang} (Member, IEEE) received the B.E. degree in automation from Shandong University, Jinan, China, in 2014, and the Ph.D. degree in robot and artificial intelligence from the Department of Electronic Engineering, The
Chinese University of Hong Kong, Hong Kong, in 2019.
During his Ph.D. study, he spent six months with the University of British Columbia, Vancouver, BC, Canada, as a Visiting Scholar. He was a Postdoctoral Fellow with the Department of Electronic Engineering, The Chinese University of Hong Kong, from 2019 to 2020. He is currently a Professor with the School of Control Science and Engineering, Shandong University. His current research interests include autonomous vehicles, active and autonomous exploration, and path planning.
\end{IEEEbiography}

\vspace{11pt}

\begin{IEEEbiography}[{\includegraphics[width=1in,height=1.25in,clip,keepaspectratio]{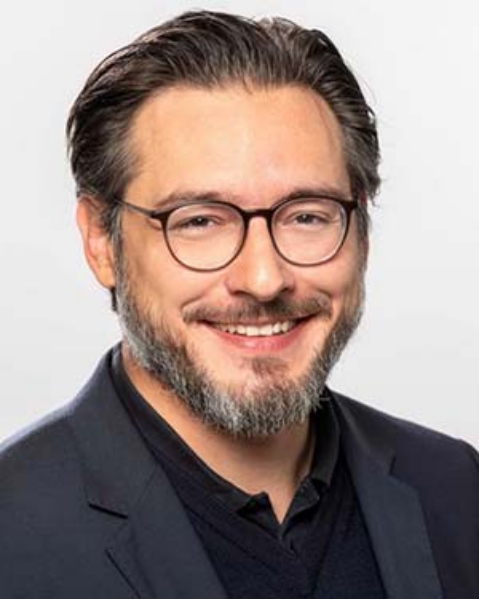}}]{Sami Haddadin} (IEEE Fellow) received the Dipl.-Ing. degree in electrical engineering in 2005, the M.Sc. degree in computer science in 2009 from the Technical University of Munich (TUM), Munich, Germany, the Honours degree in technology management in 2007 from Ludwig Maximilian University, Munich, Germany, and TUM, and the Ph.D. degree in safety in robotics from RWTH Aachen University, Aachen, Germany, in 2011. He is currently a Full Professor and the Chair with Robotics and Systems Intelligence, TUM, and the Founding Director of the Munich Institute of Robotics and Machine Intelligence (MIRMI), Munich. Dr. Haddadin was the recipient of numerous awards for his scientific work, including the George Giralt Ph.D. Award (2012), IEEE RAS Early Career Award (2015), the German President's Award for Innovation in Science and Technology (2017), and the Leibniz Prize (2019).
\end{IEEEbiography}

\begin{IEEEbiography}[{\includegraphics[width=1in,height=1.25in,clip,keepaspectratio]{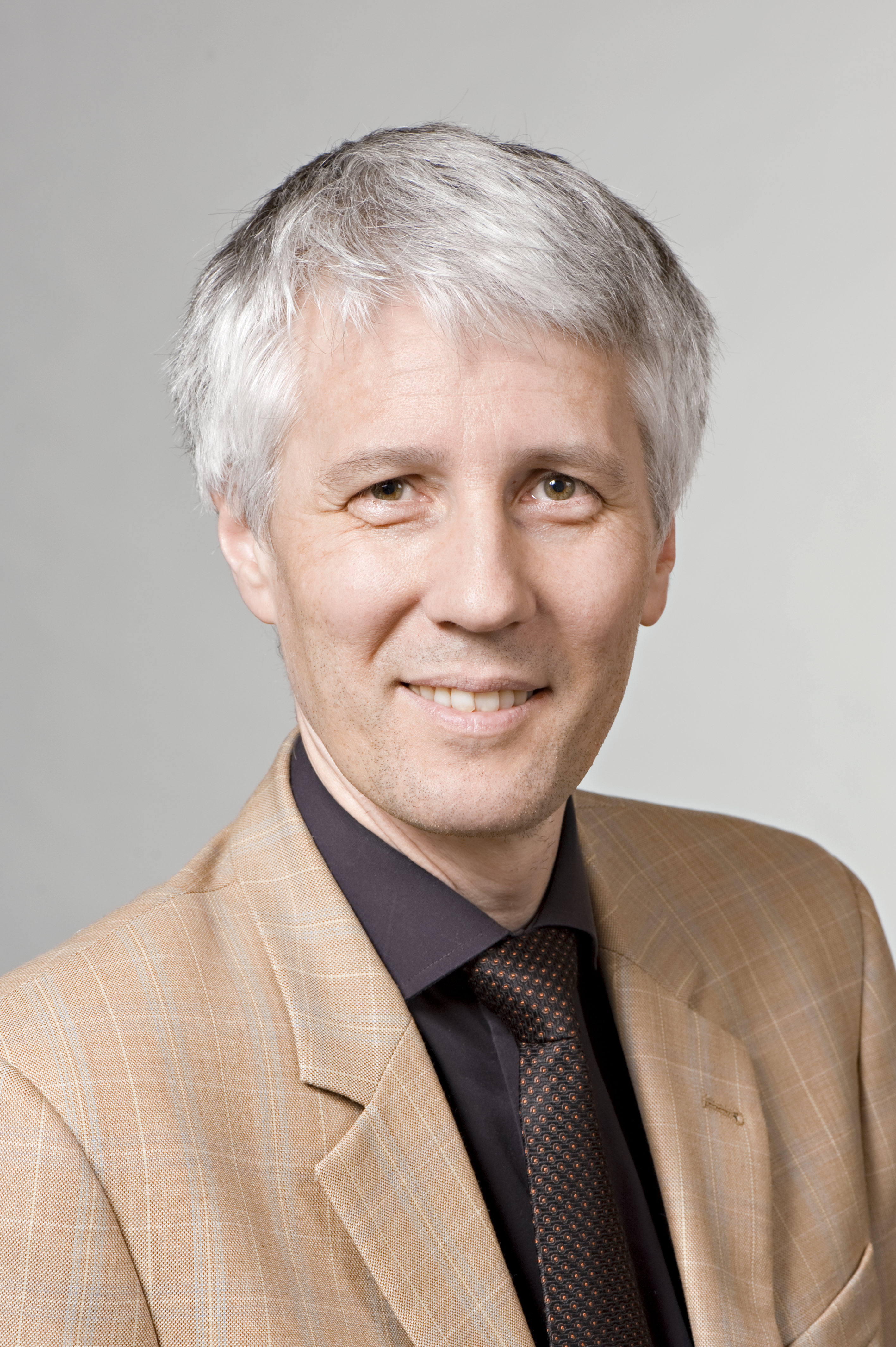}}]{Alois Knoll} (IEEE Fellow) received his diploma
(M.Sc.) degree in Electrical/Communications
Engineering from the University of Stuttgart,
Germany, in 1985 and his Ph.D. (summa cum
laude) in Computer Science from Technical University of Berlin, Germany, in 1988. He served on
the faculty of the Computer Science department
at TU Berlin until 1993. He joined the University
of Bielefeld, Germany as a full professor and
served as the director of the Technical Informatics research group until 2001. Since 2001,
he has been a professor at the Department of Informatics, Technical
University of Munich (TUM), Germany . He was also on the board of
directors of the Central Institute of Medical Technology at TUM (IMETUM). From 2004 to 2006, he was Executive Director of the Institute of
Computer Science at TUM. Between 2007 and 2009, he was a member
of the EU's highest advisory board on information technology, ISTAG,
the Information Society Technology Advisory Group, and a member
of its subgroup on Future and Emerging Technologies (FET). In this
capacity, he was actively involved in developing the concept of the EU's
FET Flagship projects. His research interests include cognitive, medical
and sensor-based robotics, multi-agent systems, data fusion, adaptive
systems, multimedia information retrieval, model-driven development of
embedded systems with applications to automotive software and electric
transportation, as well as simulation systems for robotics and traffic.

\end{IEEEbiography}

\vspace{11pt}

\begin{IEEEbiography}[{\includegraphics[width=1in,height=1.25in,clip,keepaspectratio]{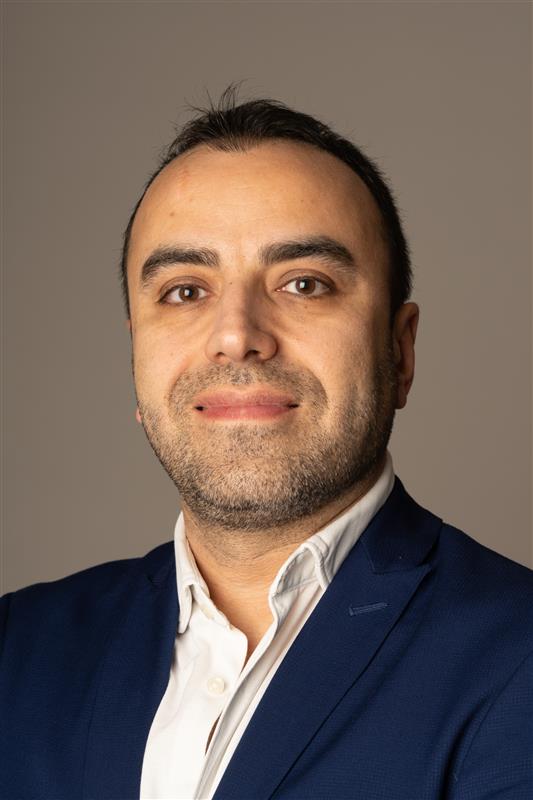}}]{Arash Ajoudani} is the director of the Human-Robot Interfaces and Interaction (HRI²) laboratory at IIT and a recipient of the ERC grants Ergo-Lean (2019) and Real-Move (2023). He has coordinated multiple Horizon 2020 and Horizon Europe projects and received numerous awards, including the IEEE RAS Early Career Award (2021) and the KUKA Innovation Award (2018). A Senior Editor of IJRR and an elected IEEE RAS AdCom member (2022-2024), his research focuses on physical human-robot interaction, mobile manipulation, and adaptive control.
\end{IEEEbiography}

\vspace{11pt}

\begin{IEEEbiography}[{\includegraphics[width=1in,height=1.25in,clip,keepaspectratio]{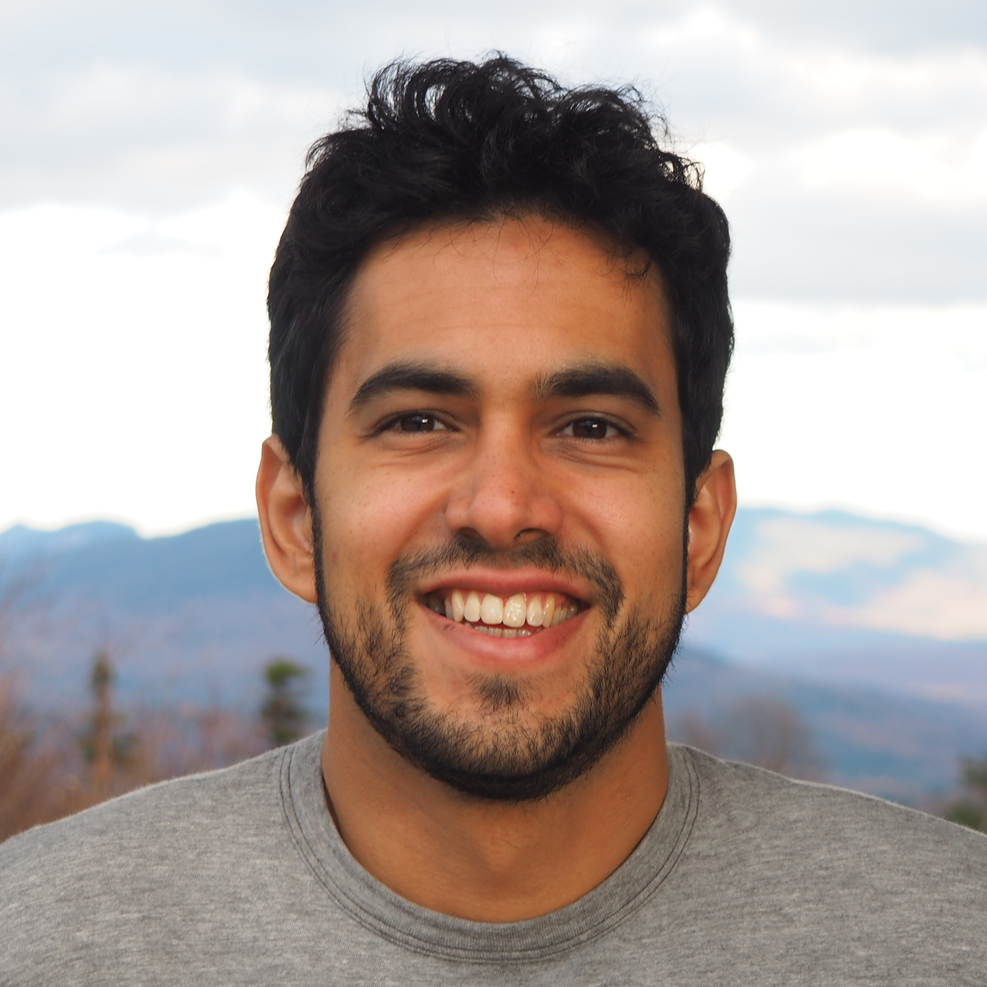}}]{Luis F.C. Figueredo} received his Bachelor’s and Master’s degrees in Electrical Engineering from the University of Brasilia, Brazil and earned his Ph.D. degree in Robotics from the same institution with an awarded PhD thesis for the biennial 2016-17. He also worked at CSAIL - MIT where he received multiple awards for robot demonstrations at venues such as IROS and ICAPS. He received the prestigious Marie Skłodowska-Curie Individual Fellowship, in 2018, for his work on biomechanics-aware human-robot interaction with AI tools acknowledged on the EU Innovation Radar, and more recently, he was also recognized within the IEEE ICRA New Generation Star Project, sponsored by NOKOV. He is currently an Assistant Professor at the University of Nottingham, UK, and an Associated Fellow at the MIRMI at TU Munich.
\end{IEEEbiography}

\vfill

\end{document}